\documentclass[letterpaper]{article} %
\usepackage[preprint]{aaai2027}  %
\usepackage[hyphens]{url}  %
\usepackage{graphicx} %
\usepackage{natbib}  %
\usepackage{caption} %
\usepackage{listings}
\usepackage{booktabs}
\usepackage{amsmath}
\usepackage{amssymb}
\usepackage{amsfonts}
\usepackage{array}
\usepackage{tabularx}
\usepackage{multirow}
\usepackage{nicefrac}
\usepackage{microtype}
\usepackage{longtable}

\newcommand{\hcode}[1]{\texttt{#1}}

\lstnewenvironment{pinkcode}{%
  \lstset{
    basicstyle=\small\ttfamily,
    frame=none,
    xleftmargin=6pt,
    xrightmargin=6pt,
    breaklines=true,
    keepspaces=true,
    columns=fullflexible,
    aboveskip=8pt,
    belowskip=8pt,
    lineskip=1pt,
    numbers=none
  }
}{}

\title{When Can One Neuron Fix Repetition Loops in LLMs?}

\author{
   Aristotelis Lazaridis \quad Aman Sharma \quad Dylan Bates \quad Brian King \\[0.5em]
   Vincent Lu \quad Jack FitzGerald \\[0.6em]
}

\affiliations{
EdgeRunner AI \\[0.3em]
   research@edgerunnerai.com
}

\begin{document}

\maketitle

\begin{abstract}
The Gemma 4 instruction-tuned models share a reproducible failure: on long factual enumeration prompts, such as listing every episode of a TV series, the 88 IAU constellations, or the 151 original Pok\'emon, they collapse into repetition, either a tight verbatim loop or a list whose entries decay onto a single answer.
These loops occur at rates as high as 87.5\% (7/8 generations) and survive prompt rewording and most sampling adjustments.
In this paper, we explore whether edits to a small number of internal model components can directly reduce this failure, rather than merely discouraging repeated tokens with a repetition penalty. This distinction matters because, as we show experimentally, repetition penalties can also distort valid repetition and degrade task performance.
To locate promising intervention targets, we combine per-layer ablation with per-neuron or routed-expert attribution, then evaluate the resulting weight edits over complete generations.
We find that these edits substantially reduce detected loops on the prompts and seeds used to select them; in the Gemma 4 E2B model, for example, one sign-inverted neuron suffices.
Across all four Gemma models, detected loops fall from 46/384 to 12/384 on frozen held-out prompts and seeds, driven mainly by the E4B and 31B models, while
separate general-purpose benchmarks show no statistically detectable regressions.
Our attribution methodology identifies useful intervention candidates, but the rankings vary across examples.
At longer generation budgets, the edits remain effective for the smaller (E2B, E4B) models, but in the larger models (26B, 31B), the remaining failures shift toward \textit{doom looping}: a non-convergent self-correction cycle over facts the model cannot recall.
In exploratory experiments on two additional model families (Qwen3.5 and LFM2.5), sparse edits also reduce repetition, providing preliminary cross-family evidence, although the strength and selectivity of the effect vary.
Overall, our results show both the promise and the limits of targeted, small-scale weight editing: it can suppress specific repetition failures and provide a training-free causal intervention, but it does not identify a universal loop circuit, guarantee clean termination, or supply missing knowledge.
\end{abstract}

\section{Introduction}

The Gemma 4 model family~\citep{gemmateam2026gemma4} exhibits a reproducible repetition failure on long factual enumeration tasks.
Prompts such as listing episodes of the series \textit{The Wire} or \textit{Firefly}, enumerating Generation I Pok\'emon, or listing constellations require the model to maintain a moderately long factual sequence while generating. Under the default sampling configuration\footnote{\texttt{temperature=0.7}, \texttt{top\_p=0.95}, \texttt{top\_k=64}, and no repetition penalty.}, the most failure-prone model--prompt pair yields loops in 7 of 8 fixed-seed generations (E2B, \texttt{constellations}, thinking disabled). The effect of reasoning mode is model- and prompt-dependent: for example, the Gemma 4 31B model with thinking enabled exhibits loops when asked to list all episodes of \textit{The Wire} (5 of 8 generations loop), whereas no generations exhibit loops with thinking disabled.

Those failures are not limited to exact string repetition: sometimes the model collapses into a tight loop, where it commits to a short phrase and repeats it until the generation budget is exhausted, while in other cases it keeps the surface structure of an enumerated list, but many entries converge to the same (repeated) answer. Both forms survive prompt rewording, inference-engine changes, and most standard sampling adjustments.

We investigate whether targeted static weight edits (i.e. \textit{weight surgery}) can reduce these repetition failures. Per-layer attribution generates candidate components, and empirical sweeps over MLP-neuron zeroing, sign inversion, amplification, and routed-expert masking select small sets whose modification reduces looping. In \texttt{gemma-4-E2B-it} (hereafter \textit{E2B}), modifying a single MLP (Multi-Layer Perceptron) neuron eliminates the loops on the original prompt-and-seed set, with a two-neuron variant reducing the associated benchmark regression. In \texttt{gemma-4-E4B-it} (\textit{E4B}) and \texttt{gemma-4-31B-it} (\textit{31B}), sets of MLP neurons are stripped (three of 430,080 neurons in E4B; 1,062 unique neurons in 31B). In \texttt{gemma-4-26B-A4B-it} (\textit{26B}), a Mixture-of-Experts (MoE) model, three layer-specific routed experts are masked out of 3840\footnote{The selected edited model variants are available on Hugging Face:\\
\url{https://huggingface.co/edgerunner-ai/gemma-4-E2B-it-noloop}\\
\url{https://huggingface.co/edgerunner-ai/gemma-4-E4B-it-noloop}\\
\url{https://huggingface.co/edgerunner-ai/gemma-4-26B-A4B-it-noloop}\\
\url{https://huggingface.co/edgerunner-ai/gemma-4-31B-it-noloop}}. Matched random edits support the selectivity of these components most clearly for E4B in both thinking modes, E2B without thinking, and 31B with thinking; they do not distinguish the selected edits for 26B or thinking-enabled E2B.

The interventions differ substantially in size: one MLP neuron in E2B, three in E4B, 1,062 in 31B, and three routed experts in 26B. This variation suggests that the number of internal components that must be modified differs across the model family. Across all four models, the edits substantially reduce or eliminate the observed loops, while separate general-purpose benchmarks show no statistically detectable regressions. Ultimately, these results establish a strong proof of concept, demonstrating that targeted interventions can eliminate repetition loops while preserving broader model performance.

This investigation is motivated by the question of whether a high-level language-model generation failure can be traced to a small set of internal components and changed through direct weight edits without retraining the model. We draw inspiration from recent mechanistic-interpretability work by \citet{kazemi2026single}, which shows that a single MLP neuron can mediate the refusal axis in safety-tuned LLMs. In the same spirit, we test whether modifying selected Gemma 4 neurons or routed experts reduces repetition. For the failure modes studied here, these modifications do reduce repetition, although the required operation differs by model and the results do not establish a complete or uniquely identified loop circuit.

The main contributions of this work are:
\begin{itemize}
  \item Diagnosis of the looping failure mode in the Gemma 4 model family.
  \item Characterization of two loop phenotypes (tight loops and soft loops), and a prompt-agnostic deterministic loop detector for both.
  \item A per-layer attribution methodology for ranking candidate MLP neurons and routed MoE expert slots associated with looping behavior.
  \item Empirical demonstrations of weight surgery feasibility across four Gemma 4 models, including a single-neuron E2B edit and sparse interventions that reduce failures in E4B, 31B, and 26B.
  \item A leakage-controlled cross-family evaluation on Qwen3.5, including a matched comparison with Final Token Preference Optimization (FTPO), a preference-training baseline designed to reduce repetition.
  \item An analysis of doom looping at extended generation budgets: a non-convergent self-correction regime that can persist in the two larger models after the primary looping failure is reduced and is consistent with a knowledge-precision gap, although our experiments do not establish its exact cause or isolate a separate, generalized loop circuit.
\end{itemize}

\section{Related Work}
\label{related}

Prior work addresses repetition through nucleus sampling
\citep{holtzman2020curious}, unlikelihood training
\citep{welleck2020unlikelihood}, and analyses of repeated-token reinforcement
and high-inflow token distributions \citep{xu2022learning,fu2021theoretical}.
Antidoom instead trains on preferences constructed at detected loop boundaries
\citep{liquidAI2026Antidoom}. These approaches modify decoding, data, or
objectives; we edit ranked weights directly and compare with Antidoom.
Mechanistic studies identify induction heads as copy mechanisms
\citep{olsson2022induction}, develop causal circuit-discovery methods
\citep{wang2022interpretability,conmy2023automated}, and isolate a
copy-suppression head \citep{mcdougall2023copy}. Although superposition
complicates neuron-level interpretation \citep{elhage2022superposition},
individual neurons and sparse feature circuits can remain interpretable and
editable \citep{gurnee2023finding,marks2024sparse}. Finally, extreme sensitivity
to a few outlier dimensions provides precedent for sparse intervention
\citep{kovaleva2021bert,puccetti2022outliers}.

\citet{hiraoka2025repetition} identify
repetition neurons and control repeated-token probabilities through activation interventions.
\citet{doan2025understanding} study their relationship to induction heads and use depth-dependent
ablation to reduce repetition while preserving in-context learning. \citet{yona2025repeated}
identify an attention-sink circuit behind repeated-token divergence and apply a targeted activation
patch. These studies establish neuron-level intervention for repetition as prior art. Our
contribution differs in baking edits into static weights, evaluating complete generated responses
across dense and mixture-of-experts Gemma 4 models, and separating intervention selection from
frozen held-out evaluation and matched controls.

For model editing, \citet{meng2022locating} introduce causal tracing and ROME for targeted MLP weight 
edits. \citet{meng2022memit} scale this to thousands of edits, and \citet{hase2023localization} find 
that localization and optimal edit layer can diverge, a result we corroborate in E2B where the layers 
identified by ablation do not coincide with the best intervention layers. \citet{turner2023steering} 
and \citet{zou2023representation} develop activation-space steering methods applied at inference time. 
Our sign-inversion edit is analogous but encoded permanently into static weights. Separately,
\citet{kazemi2026single} demonstrate that a single MLP neuron mediates the refusal axis in safety-tuned LLMs, 
inspiring our investigation of whether repetition failures are similarly editable, and
\citet{wei2024brittleness} show that safety-critical regions occupy only $\sim$3\% of parameters, a 
structural parallel to our selected edits touching 0.0007--0.085\% of a model's Feed Forward Network (FFN) neurons.

\section{Failure Phenotypes and Sampling Controls}
\label{phenotypes}

We classify the observed repetition behavior into two phenotypes. The first is a \textit{tight loop}: the model commits to a short phrase and re-emits it verbatim, repeating the same token sequence until the generation budget is exhausted. The second is a \textit{soft loop}, i.e. a list-collapse repetition: the model continues the surface structure of an enumerated list, but the contents of multiple entries collapse to the same answer.

\begin{itemize}
  \item \textbf{Tight loop.} This phenotype is most visible in the 31B and 26B models. After several hundred tokens in the thinking block, the model may commit to a single word or phrase (such as ``\texttt{S1E6: The Detail ... no.}'') and re-emit the same short token sequence until the maximum response length is reached.
  \item \textbf{Soft loop.} This phenotype is most visible in the E2B, E4B, and 26B models. The numbered list scaffold remains intact while its entries converge to one answer. On the E2B \texttt{constellations} probe without thinking, for example, the first 47 entries are mostly correct, but entries 48--88 collapse onto \texttt{Volans} (41/88 items) even as the numbering continues: ``\texttt{47. Virgo; 48. Volans; 49. Volans; \ldots; 88. Volans}.'' 
\end{itemize}

We refer to both phenotypes collectively as \textit{fast-commit loops}: failures in which the model locks onto a repeated output within the first generation pass and sustains it until the budget cap. This distinguishes them from \textit{doom looping}, a non-convergent self-correction regime that emerges at extended generation budgets, in which the model spends thousands of tokens revisiting and rephrasing the same uncertain fact without resolving it. We discuss doom looping in more detail in Section~\ref{doom_looping_sec}.

A repetition penalty of at least 1.15 can reduce repetition in several cases, but this is not a reliable solution because it can degrade unrelated generation behavior. For example, on the E4B model, increasing the repetition penalty to 1.30 breaks code generation: 22 of 30 sampled generations of one Rust code-generation prompt reach the maximum output length with progressively degraded syntax.

An example of such degraded Rust code generation is presented below:

\begin{pinkcode}
writeln!(stdout(), "\n INFERENCE STATUS:")?)?;
writeln!(stdout(), "---------------")?;
writable!(cout, Cursor::Cyan)(concat!("Active Requests: ", inf.in_flight_requests))?;
writable!(cout, Cursor::Magenta)(concat!("Avg Throughput: ", nf!(inf.avg_tokens_per_second, ".1")))?;
writable!(cout, Cursor::White)(concat!("Latest ID Found: ", &(nf!(inf
\end{pinkcode}

In this example, the model has drifted into invented syntax: \texttt{writable!} is not a Rust macro; it is a hallucinated alternative that arose after \texttt{writeln!} was penalized for having appeared previously. The same effect is visible in refusal language: at \texttt{repetition\_penalty=1.30}, ``\textit{I cannot help with that}'' becomes ``\textit{I cannot help with thus}'' or ``\textit{I cannot assist with such}'', because the standard phrasing is penalized after prior use. At \texttt{repetition\_penalty=1.15} and at the baseline value of 1.00, the code generation is unaffected in our tests.

This side effect was not observed on the 31B model in our Rust tests, though we expect the same mechanism to manifest on other prompts or models where idiomatic token margins are smaller. More broadly, the side effects of a repetition penalty are workload- and model-dependent, which makes it an unreliable general solution. The selected weight edits act on attributed internal candidates and operate at \texttt{repetition\_penalty=1.0}, avoiding this tradeoff entirely. The full quantitative results are reported in Supplementary Sec.~G.

\section{Methodology - Experimental Setup}
\label{method}

This section describes our prompt suite for triggering loops, the loop detection mechanism, the activation-based localization procedure, and our evaluation benchmark protocol used throughout the paper.

We use eight \textit{enumeration probes}: \texttt{firefly\_list},
\texttt{wire\_episodes}, \texttt{pokemon\_gen1}, \texttt{mcu\_films},
\texttt{noble\_gases}, \texttt{constellations}, \texttt{us\_presidents}, and
\texttt{eu\_member\_states}. These factual-list prompts span television episodes,
scientific categories, and other closed or approximately closed sets; full prompt
text, targets, and failure-mode roles are given in Supplementary
Sec.~A.

For each model variant, the initial loop evaluation consists of these 8 probes $\times$ 8 fixed seeds (0-7) $\times$ 2 thinking modes (thinking on or off), for 128 generations per model variant. We generate up to 1536 new tokens with \texttt{temperature=0.7}, \texttt{top\_p=0.95}, \texttt{top\_k=64}, and no repetition penalty. Unless otherwise specified, loop rates are reported per thinking mode over 64 generations, since the loops are primarily triggered during thinking. The fixed seeds make each evaluation reproducible within its pinned inference implementation, while the rates estimate behavior under sampling and may differ for other prompts, seeds, or implementations. Because this same prompt-and-seed set was used to select the interventions, we treat these as development-set results and evaluate generalization separately on held-out data.

To classify outputs, we created a deterministic, prompt-agnostic detector. A tight loop requires an exact 5--40-token period repeated at least three times at the generation tail. A soft loop requires at least four consecutive lines with identical normalized content after stripping list markers, lowercasing, and normalizing whitespace.\footnote{We also implemented an auxiliary exact-character check for an 8--120-character unit repeated contiguously at least four times. It is not a separate paper phenotype and did not determine any reported initial or long-budget result; Supplementary Sec.~E gives an example.} An output classified as non-looping may still be incomplete or factually incorrect; the verdict means only that these repetition patterns were not detected. On a stratified 81-output blinded human audit, the detector has precision 0.878, recall 0.818, and F1 0.847 (Supplementary Sec.~B).

For regression checks, we report IFEval~\citep{zhou2023instruction},
TruthfulQA~\citep{lin2022truthfulqa}, ARC-Easy and
ARC-Challenge~\citep{clark2018arc}, and GSM8K~\citep{cobbe2021gsm8k}. Each benchmark is run in both thinking
modes with a separate deterministic benchmark configuration: \texttt{temperature=0.0, top\_p=0.95, max\_output\_tokens=8192} and no repetition penalty.
We excluded MMLU~\citep{hendrycks2021mmlu} because its task-level short answer cap truncates these models'
prose-first responses and yields an invalid comparison.

After freezing the selected interventions, we evaluate behavior on unseen
seeds 8--15 and six prompts not used for intervention selection: paraphrases of
the \texttt{firefly}, \texttt{wire}, \texttt{constellations}, and
\texttt{pokemon} triggers, plus two new behavior-preservation probes that request
all U.S. states and all chemical elements. We use the same 1536-token decoding
setup and test both thinking modes
($6$ prompts $\times$ $8$ seeds $\times$ $2$ modes $\times$ baseline/selected
for each model; 768 generations total). Our main held-out outcome is the fraction
of generations classified as loops by the detector. We also measure natural EOS,
budget exhaustion, unique and duplicate list items, and, for the U.S.-states
probe, coverage of the 50-item gold list. These additional measures test whether
an edit truncates or degrades an answer; they are not expected to show that an
edit adds missing knowledge.
For selectivity, we additionally freeze five matched random interventions per
model. Each uses the selected edit's layer, operation, and unit count while
excluding the selected units. For 26B, each random mask also matches the selected
mask's layer allocation: two routed experts at L21 and one at L22. Every random
intervention is evaluated on the same held-out suite.

For each model, we capture one thinking-enabled rollout that reliably loops and
identify its exact token-period window. We compare generated positions
\texttt{[200, start$-$50]} (pre-loop reasoning, excluding the transition) with
\texttt{[start$+$50, $N$]} (the established loop). The capture prompts are
\texttt{firefly\_list} for 31B/E4B, \texttt{constellations} for E2B, and
\texttt{wire\_episodes} for 26B.

The static edits below target the fast-commit loop mechanism (defined in Section~\ref{phenotypes}), which is distinct from the doom looping (discussed later in Section~\ref{doom_looping_sec}).

The first localization step is a \textit{per-layer ablation sweep}, i.e. testing every layer in turn. For each layer $\ell \in \{0 \ldots L{-}1\}$, we set the output of either attention or MLP to zero on the captured rollout and measure how the probability of the loop token under the next-token distribution changes. This identifies candidate anti-loop layers: depths where removing a component strongly changes the probability of the next loop token. These candidate layers guide attribution, but as we observed they do not by themselves determine the final intervention layer.

For E2B, this procedure reveals a dissociation: the per-layer ablation identifies L13--L15 as the site of the strongest loop-token signal (where the current loop token is most directly written into the residual stream), but interventions at those layers did not produce the best behavioral outcome. The selected edit acts earlier, at L10/L12, and was selected after running full-generation loop-rate sweeps over candidate layers. Single-token ablation can reveal where the loop token is committed, while full-generation sweeps are needed to find where a static edit changes whether the trajectory enters the loop at all. This is a behavior-level instance of a disconnect, which is in line with the findings on factual-knowledge editing by \citet{hase2023localization}: the location where a behavior is localized need not be the location where editing it works best. We give the detailed per-layer evidence for E2B in Supplementary Sec.~H.

On the original captured rollouts, the ablation sweeps rank L36 for 31B, L18 for E4B, and a broad L21--L26 signal for 26B; the selected expert masks act at L21--L22. For E2B, the strongest single-token signal appears at L13--L15, whereas the effective intervention acts earlier at L10/L12. Full plots and the layer summary are reported in Supplementary Sec.~C.

For E2B, E4B, and 31B, a neuron is an intermediate-dimension index in the
GeGLU MLP~\citep{shazeer2020glu,dauphin2017language}; interventions modify its
\texttt{down\_proj} column. We rank all neurons in a candidate layer by the
gradient-times-activation score
\[
\Delta_n^{(\ell)}
\;=\;-\sum_{p}a_n^{(\ell)}(p)
\frac{\partial \log p_{\theta}(t\mid x_{<p})}{\partial a_n^{(\ell)}(p)}.
\]
Positive scores identify anti-loop units (candidates for amplification); negative
scores identify pro-loop units (candidates for suppression, zeroing, or inversion).
We confirm top-$K$ candidates with exact weight edits and full-generation sweeps.
Supplementary Sec.~C gives the full GeGLU definition,
counterfactual interpretation, and layer-level evidence.

However, attribution is not stable across looping rollouts. Across six reproducible
captures per model, the strongest ablation layer never matched the originally
selected layer. Within the same prompt, neuron-rank Spearman correlation and
top-100 overlap average 0.056/0.097 for E4B and 0.122/0.205 for 31B. For 31B,
two additional detector-positive generations from the vLLM selection run did not
loop when recaptured through the HF attribution path, so they were excluded
before any attribution scores were computed. We therefore use attribution only
to generate candidates; behavioral sweeps and held-out comparisons support the
final edits.

For the Gemma 4 26B MoE model, the analogous object is not an MLP neuron but a routed expert
slot. We apply the same pre-loop vs loop contrast to router-selection probabilities
for each \texttt{(layer, expert)} pair. This produces a small set of expert positions that
are selected far more often once the model enters the loop.

The following section evaluates static weight-space interventions derived from these attribution and sweep results.

\section{Intervention Search and Results}

We evaluate four classes of static weight interventions on the top-$K$ selected units:

\begin{itemize}
  \item \textit{Stripping}: zero the selected \texttt{down\_proj} columns, removing those neurons' write contribution to the residual stream. Analogously, in the MoE case (26B), zero the post-router scale of the selected routed experts.
  \item \textit{Amplification}: scale the selected units by a factor $\alpha > 1$.
  \item \textit{Suppression}: scale the selected units by a factor $0 < \beta < 1$.
  \item \textit{Sign inversion}: scale a selected MLP column by $\alpha < 0$, converting a pro-loop direction into an active anti-loop direction.
\end{itemize}

For the 31B and E4B models, stripping the selected MLP neurons produced the lowest loop rates without measurable benchmark degradation. E2B required a different intervention direction: stripping the most pro-loop L10 neurons reduced failures but left residual soft-loop failures, whereas sign inversion of the dominant L10 pro-loop neuron eliminated the failures. For the 26B model, shared-MLP interventions were not appropriate because the shared MLP accounts for only about $27\%$ of FFN compute. Routed experts carry the remaining FFN computation, so expert masking was selected as the intervention method instead.

\subsection{Selected interventions per model}
\label{dense_results}

The dense-model edits are: 1,062 unique L36 neurons in 31B (0.082\% of FFN
neurons), three L18 neurons in E4B (0.0007\%), and two neurons in E2B
(0.0009\%): L10:3513 is sign-inverted with $\alpha=-0.8$ and L12:3838 is
amplified with $\alpha=3.0$. Table~\ref{tab:selected_variants} reports the
loop and benchmark outcomes on the original prompt-and-seed set.

\subsection{E2B: sign inversion of a pro-loop neuron}

E2B differs from E4B and 31B in two ways. First, its baseline failures are concentrated almost entirely in the \texttt{constellations} probe: 6/64 failures with thinking enabled and 7/64 with thinking disabled. Second, all baseline failures are soft loops rather than tight loops. Zeroing the three strongest pro-loop L10 neurons reduced failures from 13/128 to 4/128. The four residual seeds still locked onto constellation names- different ones in different seeds, meaning other L10 neurons can sustain the same failure when the top three are removed.

Sign inversion resolves this zeroing-only limitation. A one-neuron variant that multiplies the L10 neuron 
3513 by $\alpha=-1$ reaches 0/128 failures, but has a worse benchmark tail of about $-2.6$ pp. The 
selected two-neuron variant tunes this sign inversion to $\alpha=-0.8$ and adds an L12 anti-loop 
amplification at $\alpha=3.0$, giving 0/128 failures with worst regression $-0.74$ pp and average absolute 
benchmark delta 0.56 pp. Supplementary Sec.~H reports the complete Pareto frontier and sweep.

One interpretation is that zeroing permits other L10 neurons to sustain the
same list-collapse trajectory, whereas sign inversion changes that trajectory;
the experiments do not establish a unique underlying circuit. L12 amplification
was retained because it gave the smallest measured benchmark regression among
the tested variants.

\subsection{Probe-combination lesson for 31B}

For 31B, combining 1000 Firefly-ranked and 100 Wire-ranked candidates eliminates
all thinking-enabled development-grid loops, whereas a shared summed ranking
leaves 1/64. This prompt-specific behavior motivates using separate
prompt-specific candidate sets rather than interpreting them as evidence for
distinct circuits. Supplementary Sec.~D.1 reports all strategies.

\subsection{MoE model: masking routed experts}
\label{moe_masking}

The 26B model has 30 layers with 128 routed expert slots per layer. Shared-MLP
stripping caused large regressions (e.g. -34\,pp for GSM8K at $K{=}100$ with thinking enabled), so we instead rank \texttt{(layer, expert)}
slots by their change in routing probability between pre-loop and loop tokens.
The selected static mask zeroes L21:E47, L21:E98, and L22:E47 (3/3840 slots).
With thinking enabled, it reduces loops from 6/64 to 2/64 across the full
eight-probe development grid; within its eight-seed \texttt{wire\_episodes}
subset, the reduction is 5/8 to 0/8. Its held-out effect is inconclusive and does
not separate from matched random masks. Supplementary Sec.~D.2 gives the
architecture, rankings, and complete mask sweep.

\section{Results}
\label{results_sec}

The selected variants are the smallest interventions found in our exploratory
search on the original prompt-and-seed set that reduce the main loop phenotype
without large measured benchmark changes. Table~\ref{tab:selected_variants} summarizes their
loop counts on that evaluation, benchmark deltas, and long-budget behavior; exact internal
identifiers are reported in Supplementary Sec.~I.

On the development evaluation used to select the interventions, the edits
eliminate or substantially reduce the observed loop behavior across all four
models (Table~\ref{tab:selected_variants}). In the
thinking-enabled mode, E2B, E4B, and 31B reach zero detected loops; small residuals
remain for E4B and 31B with thinking disabled and for 26B in both modes.

\begin{table}[t]
\centering\footnotesize
\begin{tabularx}{\columnwidth}{@{}l X c@{}}
\toprule
\textbf{model} & \textbf{edit (units; benchmark $\Delta$ pp)} &
\textbf{loops on/off} \\
\midrule
E2B & L10 inversion + L12 amplification
  (2 neurons; $[-0.74,+1.71]$) &
$6\!\rightarrow\!0$ / $7\!\rightarrow\!0$ \\
E4B & L18 MLP strip
  (3 neurons; $[-0.86,+2.03]$) &
$6\!\rightarrow\!0$ / $8\!\rightarrow\!2$ \\
31B & Probe-combined L36 MLP strip
  (1,062 unique neurons; $[-0.37,+0.80]$) &
$13\!\rightarrow\!0$ / $1\!\rightarrow\!1$ \\
26B & L21/L22 routed-expert mask
  (3 experts; $[-0.86,+1.59]$) &
$6\!\rightarrow\!2$ / $4\!\rightarrow\!1$ \\
\bottomrule
\end{tabularx}
\caption{Selected interventions. Loops are baseline$\rightarrow$selected over 64 development generations per thinking mode (on/off). Benchmark deltas exclude MMLU.}
\label{tab:selected_variants}
\end{table}

\subsection{Held-Out Generalization and Task Quality}

Across 384 held-out baseline and 384 paired selected-edit generations, detected
loops fall from 46 to 12. The effect is heterogeneous: E4B falls from 18/96 to
2/96 (paired reduction 16.7\,pp, 95\% bootstrap CI [8.3, 25.0]) and 31B from
13/96 to 3/96 (10.4\,pp [4.2, 16.7]); E2B falls from 6/96 to 1/96
(5.2\,pp [0.0, 10.4]), while 26B falls from 9/96 to 6/96
(3.1\,pp [$-3.1$, 9.4]). Thus the strongest held-out evidence is for E4B and
31B; the 26B effect is not distinguishable from zero at this sample size.

Five matched random interventions per model clarify selectivity. The selected
edit has a lower loop rate than every draw for E4B in both modes, E2B without
thinking, and 31B with thinking. It lies inside the random range for 26B and
thinking-enabled E2B, while 31B without thinking is at a floor. Therefore, when
the selected edit does not outperform the random controls, we treat attribution
as a way to generate intervention candidates rather than as evidence that it
has identified a specific loop circuit (Supplementary Sec.~B).

\begin{figure*}[t]
  \centering
  \includegraphics[width=0.80\textwidth]{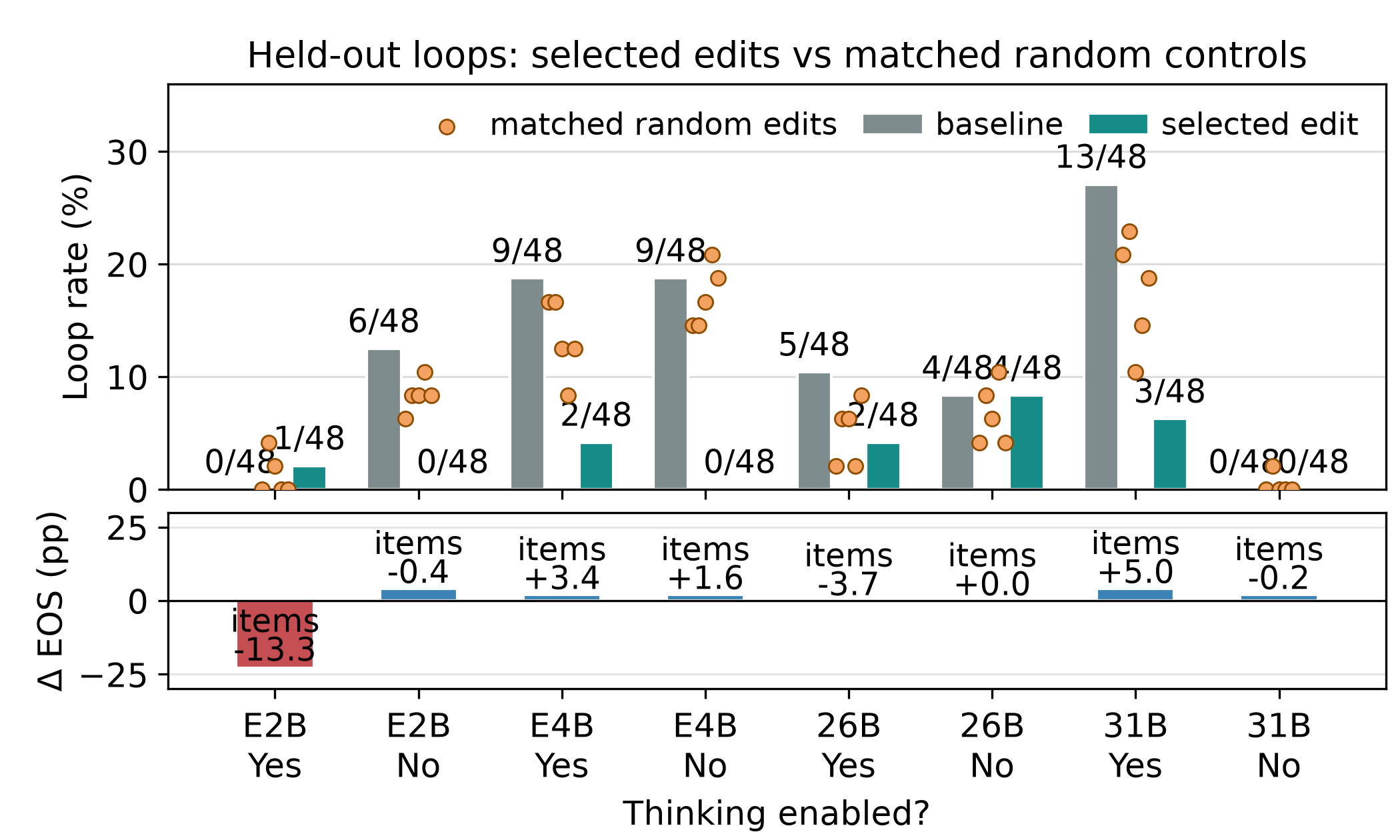}
  \caption{Held-out baseline and selected loop rates, with five matched random
  interventions for each model and thinking mode. The lower panel shows
  selected-minus-baseline natural-EOS changes and labels the corresponding
  change in unique parsed list items. Each bar contains 48 generations.}
  \label{fig:heldout_controls}
\end{figure*}

Because this test uses only six prompts, its 384 generations are not fully
independent: runs from the same prompt are closely related. We therefore repeat
the analysis by resampling prompts and then seeds within each prompt. All
detected loops occur on the four paraphrased triggers, where loops fall from
46/256 to 12/256. Neither the baseline models nor the edited models loop on the two new
controls (0/128 for both). Across all models, the estimated reduction is
8.9 percentage points (95\% CI [1.8, 18.0]). However, the prompt-level interval
for each model includes zero. These results support loop reduction on the
tested prompts, but do not establish broad generalization to other tasks
(Supplementary Sec.~B).

However, the absence of detected repetition does not imply completeness or
correctness. Across held-out closed-list prompts, coverage is essentially
unchanged for E4B (0.882$\rightarrow$0.893), 31B
(0.908$\rightarrow$0.911), and 26B (0.992$\rightarrow$0.990), with similar
precision and natural-EOS rates. E2B is the exception: coverage falls from
0.795 to 0.695 and natural-EOS from 0.906 to 0.781, concentrated on the longer
chemical-elements and Pok\'emon prompts. We therefore treat E2B as a
loop-completion trade-off rather than an unqualified fix.

To test the headline single-neuron result beyond the data used to select it, we
also evaluate that edit separately on the E2B held-out set. It manages to eliminate all 13
loops across the 128-generation development grid, but on the 96 held-out
generations it reduces loops only from 6 to 4, resolving five baseline loops
while introducing three new ones (exact McNemar $p=0.727$). Natural termination
also falls from 90/96 to 81/96. Thus the original result is correct for the
selection grid, but the one-neuron edit does not show comparably strong
generalization. The two-neuron E2B variant has the stronger held-out loop result,
but it still exhibits the completion trade-off described above.

On five development benchmarks in both thinking modes, measured deltas remain
within $[-0.74,+1.71]$ pp for E2B, $[-0.86,+2.03]$ pp for E4B,
$[-0.37,+0.80]$ pp for 31B, and $[-0.86,+1.59]$ pp for 26B
using Inspect-AI~\cite{UK_AI_Security_Institute_Inspect_AI_Framework_2024}.

We also run three untouched, no-thinking benchmarks that were not used to select
variants: HellaSwag~\citep{zellers2019hellaswag},
WinoGrande~\citep{sakaguchi2020winogrande}, and
BoolQ~\citep{clark2019boolq}; 500 examples each. Selected-minus-baseline
changes range from $-3.4$ to $+0.4$\,pp for E2B, $+0.6$ to $+4.0$\,pp for E4B,
$0.0$ to $+1.4$\,pp for 26B, and $-0.8$ to $+0.2$\,pp for 31B. Conservative
95\% intervals based on the two reported standard errors include zero in all
12 model--benchmark comparisons (Supplementary Sec.~B), so none shows a
statistically detectable regression.

\subsection{Cross-Family and Training Baselines}

To test whether the approach transfers beyond Gemma, we screened prompts from
Antidoom's released training mixture~\citep{liquidAI2026Antidoom} with
Qwen3.5-4B~\citep{qwen35modelcard}, then froze a fresh 400-prompt Qwen
confirmation set after trigger discovery. We evaluated every variant on the
same inputs with both our detector and Antidoom's detector. A 100-neuron L12 sign-inversion edit
reduces positives from 181 to 104 under Antidoom's
repeated-span detector (111 resolved, 34 introduced; exact McNemar
$p<10^{-10}$) and from 74 to 42 under our detector (56 resolved, 24 introduced;
$p=0.00045$). These results show that sparse weight editing can also reduce
repetition outside the Gemma family. However, they do not establish a universal
loop circuit: the two detectors capture different forms of repetition,
attribution rankings remain unstable, and the edit introduces some new loops.
The initial 32-prompt grid selected \texttt{L12-K30-zero} by its frozen
tie-break; the remaining 108 development prompts refined the intervention to
\texttt{L12-K100-invert}, which is the edit reported on the disjoint
400-prompt confirmation set.

For comparison, we followed the procedure proposed in
Antidoom~\citep{liquidAI2026Antidoom} and trained an FTPO baseline for
Qwen3.5-4B using a disjoint pool of 20,000 generated preference pairs. Its
early-stopped checkpoint yields 0/400 positives (i.e. loops) under both detectors, while
preserving accuracy across the seven capability benchmarks. However, a
full-epoch checkpoint is worse on both axes: it leaves 23/400 and 26/400
positives under our and Antidoom's detectors, respectively, and sharply degrades
six of the seven capability scores. On E4B, the same Antidoom procedure yielded
only 2,080 generated pairs, and matched training gives no net held-out improvement
(18/96 to 18/96; four resolved and four introduced). On Qwen, FTPO removes more
detected loops when preference-data and training resources are available.

To determine whether these outputs actually finish, we also examine vLLM's stop
reason. Of the 181 baseline outputs flagged by Antidoom, the sparse edit removes
detected repetition in 111; 57 then terminate naturally, while 54 still reach
the token limit. FTPO removes detected repetition in all 181; 68 terminate
naturally, while 113 still reach the token limit.
Across all 400 prompts, natural stopping rises from 95 at baseline to 166 with
the sparse edit and 189 with FTPO. Thus, FTPO suppresses more detected
repetition, but its 0/400 loop count does not imply that all outputs complete
successfully. Once identified, the sparse edit is compact, requires no
preference-pair generation or training, and enables component-level
causal tests.

A smaller LFM2.5-1.2B-Thinking experiment~\citep{liquidAI2026thinking} provides a second cross-family
feasibility result: a 100-neuron edit reduces held-out loops from 4/9 to 1/9.
Because the sample is small and one of five matched random edits ties this
result, we do not interpret it as evidence that the selected units are uniquely
loop-specific (Supplementary Sec.~I).

\section{Discussion}
\label{discussion_sec}

\subsection{Doom Looping}
\label{doom_looping_sec}

At 4k--8k-token budgets, E2B and E4B retain the reduction in fast-commit
loops. In 26B and 31B, however, difficult factual prompts often produce
\textit{doom looping}: sustained self-correction that either settles into
verbatim repetition or repeatedly restarts until budget exhaustion. At
\texttt{rep\_pen=1.0}, the edits reduce the tight-loop surface form but mostly
shift failures toward non-verbatim self-correction. At \texttt{rep\_pen=1.15} natural termination is improved, but the regime is not eliminated (Supplementary Sec.~F).

For example, a 31B generation can spend thousands of tokens revisiting one
episode title: ``\textit{S1E6: The Detail \ldots no \ldots I am looping. Let me
just list the titles correctly \ldots}''. This resembles the circular-reasoning
failure studied by \citet{duan2026circular}: the model repeatedly treats its own
uncertain revisions as new evidence without converging. Some runs eventually
lock onto a short correction template, which the detector flags; others keep
rephrasing statements such as ``\textit{I must be hallucinating}'' and
``\textit{Let me start over}'' until the budget is exhausted. The latter are
not detector-positive exact loops, which is why loop counts alone understate
non-convergence.

Residual doom looping is consistent with a factual-precision problem:
suppressing repetition cannot supply a missing fact. The pooled results show
that the edit primarily removes verbatim lock-in, while the remaining failures
often become endless self-correction. A targeted 26B self-correction mask
removed late tight loops but introduced failures on clean prompts, suggesting
entanglement with general verification (Supplementary Sec.~F).

\subsection{Limitations and Future Work}
\label{limitations_sec}

This study is exploratory rather than an exhaustive search over intervention types,
layers, and unit counts.
Sparse edits reduce loops in Gemma 4, Qwen3.5, and LFM2.5, providing initial evidence that the intervention strategy transfers
across architectures. However, the selected
components differ and attribution rankings are unstable; we therefore do not
claim that these models share a
single family-independent loop circuit. Moreover, intervention selection and the initial
reported outcomes use the same prompts and seeds. We therefore treat those
results as development evidence and use the separate held-out evaluation to
assess behavior on unseen seeds and related prompts. Because four of its six
prompts are paraphrases of development triggers, task-level generalization
remains uncertain even when generation-level paired effects are strong.
Detector-negative outputs may still be incomplete or incorrect, and finite
capability tests cannot exclude effects on untested behaviors. The Rust check
uses one prompt, and the Qwen confirmation-set detector comparison does not
include independent human labels. Future work should test reversible or
detector-gated/inference-time edits and broader capability.

\section{Conclusion}

In this work, we presented a reproducible enumeration-loop failure in the Gemma 4 family of models and 
showed that
small static weight edits can suppress it with modest measured benchmark changes. Our
successful development-grid interventions that we presented are model-specific: one sign-inverted
neuron suffices in E2B (with a two-neuron variant giving the best trade-off),
three stripped neurons in E4B, a larger probe-combined stripping in 31B, and three masked
experts in 26B.
Matched controls support those interventions most clearly for E4B and
thinking-enabled 31B, while they come with a completion trade-off for E2B, and are inconclusive for 
the 26B. We also found that attribution rankings can vary substantially
across looping captures, making behavioral validation essential. Additionally, at long
budgets, doom looping can persist in the larger models when factual
uncertainty prevents convergence.
In order to test the generalization of our approach, we also conducted preliminary cross-family experiments on 
Qwen3.5 and LFM2.5, showing that training-free sparse editing can transfer beyond Gemma 4.
Following the Antidoom approach, we found that early-stopped FTPO suppresses more 
detector-positive Qwen loops, although most resolved cases still exhaust the generation budget. The two
methods therefore offer distinct intervention routes: sparse weight surgery
provides a compact, training-free intervention and a component-level causal probe,
whereas FTPO uses generated preference data and training to alter the
model's loop behavior.

\bibliography{refs}

\clearpage
\onecolumn
\begin{center}
  {\LARGE\bfseries Supplementary Material\par}
  \vspace{0.5em}
\end{center}
\vspace{1em}

\appendix

\section{Evaluation Probes}
\label{appendix:probes}

All experiments use the following 8 enumeration probes.
Each probe is run with 8 random seeds, giving 64 generations for each
model, variant, and thinking-mode combination in the initial 1.5k-token evaluation.
Table~\ref{tab:probes} lists the probes along with their enumeration target and role in the
evaluation. The full per-generation table in Appendix~\ref{appendix:full_enum} uses abbreviated probe
IDs to fit the seed columns: \texttt{firefly} = \texttt{firefly\_list},
\texttt{noble\_gas} = \texttt{noble\_gases}, \texttt{us\_pres} = \texttt{us\_presidents},
\texttt{eu\_states} = \texttt{eu\_member\_states}, \texttt{mcu} = \texttt{mcu\_films}; the
remaining three IDs (\texttt{constellations}, \texttt{pokemon\_gen1}, \texttt{wire\_episodes})
are unchanged.

\begin{table}[htbp]
\centering\small
\renewcommand{\arraystretch}{1.35}
\begin{tabularx}{\textwidth}{@{}p{3.1cm}X p{2.5cm}p{3.8cm}@{}}
\toprule
\textbf{Probe ID} & \textbf{User prompt} & \textbf{Target} & \textbf{Role in evaluation} \\
\midrule
\texttt{constellations}
  & \textit{List all 88 modern IAU constellations alphabetically. Just the Latin names.}
  & 88 Latin names (closed set)
  & \textbf{Primary E2B failure prompt} (soft loops). Control for all other models. \\[3pt]
\texttt{eu\_\allowbreak member\_\allowbreak states}
  & \textit{List every EU member state in alphabetical order, including the year each joined.}
  & 27 names + accession year
  & Control (rarely triggers loops in any model). \\[3pt]
\texttt{firefly\_list}
  & \textit{List all of the episodes of the TV series Firefly.}
  & 14 episode titles (including unaired)
  & \textbf{Primary E4B failure prompt}; also triggers soft loops in E2B.
    Factually uncertain: the model frequently hallucinates or repeats ``The Message''. \\[3pt]
\texttt{mcu\_films}
  & \textit{List every Marvel Cinematic Universe film released between 2008 and 2023, in chronological release order.}
  & $\approx$30 titles
  & Control. \\[3pt]
\texttt{noble\_gases}
  & \textit{List all the noble gases.}
  & 6--7 elements
  & Control (trivial enumeration; included to catch extreme regressions). \\[3pt]
\texttt{pokemon\_gen1}
  & \textit{List all 151 original Pokémon in order from \#001 (Bulbasaur) to \#151 (Mew). Just the numbers and names.}
  & 151 numbered names
  & Stress test: long enumeration that occasionally surfaces soft loops. \\[3pt]
\texttt{us\_presidents}
  & \textit{List all U.S.\ Presidents in chronological order, numbered from 1 to 47. Just the numbers, names, and years served.}
  & 47 entries
  & Stress test: long enumeration with knowledge ambiguity in recent entries. \\[3pt]
\texttt{wire\_episodes}
  & \textit{List every episode of HBO's The Wire across all five seasons. Use the format ``Season X Episode Y: Title''.}
  & 60 episode titles
  & \textbf{Canonical doom-loop prompt}. The only probe where 31B reliably enters
    tight loops (model never emits \texttt{<turn|>}).
    Also the worst failure prompt for 26B and one of the three prompts in the
    long-budget doom sweep. \\
\bottomrule
\end{tabularx}
\caption{Enumeration probes used throughout all sweeps.
  Bold entries mark the primary failure prompts for each model.
  ``Control'' probes produce near-zero baseline loop rates and are used to
  detect regressions introduced by a candidate intervention.}
\label{tab:probes}
\end{table}

\FloatBarrier

\section{Held-Out Evaluation, Controls, and Detector Audit}
\label{appendix:heldout}

\subsection{Frozen Protocol}

We froze the four selected interventions before running the held-out suite. The
suite uses unseen sampling seeds 8--15 and six prompts: four rephrasings of
\texttt{firefly\_list}, \texttt{wire\_episodes}, \texttt{constellations}, and
\texttt{pokemon\_gen1}, plus new controls that request all 50 U.S. states and
all 118 chemical elements. Each prompt is run with thinking enabled and disabled
at \texttt{max\_new\_tokens=1536}, \texttt{temperature=0.7},
\texttt{top\_p=0.95}, \texttt{top\_k=64}, and no repetition penalty. This gives
$6\times8\times2=96$ generations for each model--variant pair and 768 total
baseline/selected generations. Inputs, scripts, and selected checkpoint paths
were hash-locked in the submission manifest before outputs were inspected.

The primary outcome is detector-positive loop incidence. Secondary outcomes are
loop phenotype, natural EOS, and budget exhaustion. We parse numbered list items
after removing Gemma's thinking channel, then report unique items and duplicate
rate as behavior-preservation diagnostics. For the U.S.-states control, exact
matching against a 50-item key additionally measures coverage. These diagnostics
ask whether the edit removes facts the baseline already produced or merely
truncates output; they do not require the model to solve every factual list.

\subsection{Held-Out Outcomes}

Table~\ref{tab:heldout_outcomes} reports all model--mode aggregates. Across the
four models, loops fall from 46/384 baseline generations to 12/384 selected-edit
generations. Model-level paired bootstrap reductions (10,000 resamples) are
16.7\,pp [8.3, 25.0] for E4B, 10.4\,pp [4.2, 16.7] for 31B, 5.2\,pp
[0.0, 10.4] for E2B, and 3.1\,pp [$-3.1$, 9.4] for 26B. The paired exact
sign-test values are respectively $p=0.0001$, $0.0020$, $0.1250$, and
$0.5078$. These tests are descriptive and are not corrected across models.

The generation-level intervals above condition on the six fixed prompts. As a
sensitivity analysis, we additionally resample prompts within each model and
then paired seeds within prompt (50,000 draws; seed 27012027). All detected
loops occur in the four paraphrases
(46/256$\rightarrow$12/256; reduction 13.3\,pp, 95\% cluster-bootstrap CI
[3.1, 25.8]); the two new controls remain at
0/128$\rightarrow$0/128. Across all six prompts, the reduction is 8.9\,pp
[1.8, 18.0]. Model-specific intervals are E2B
[$-1.0$, 15.6], E4B [$-2.1$, 49.0], 26B [$-2.1$, 11.5], and 31B
[0.0, 25.0]\,pp. With only six fixed prompts, this analysis is descriptive and
we do not estimate a mixed-effects model; it shows that task-level uncertainty
is wider than generation-level resampling suggests.

\begin{table}[htbp]
\centering\small
\begin{tabular}{llrrr}
\toprule
\textbf{model} & \textbf{thinking} & \textbf{loops B$\rightarrow$S} &
\textbf{natural EOS B$\rightarrow$S} & \textbf{$\Delta$ unique}\\
\midrule
E2B & enabled  & $0\rightarrow1$  & $44\rightarrow33$ & $-13.31$\\
E2B & disabled & $6\rightarrow0$  & $46\rightarrow48$ & $-0.38$\\
E4B & enabled  & $9\rightarrow2$  & $38\rightarrow39$ & $+3.44$\\
E4B & disabled & $9\rightarrow0$  & $47\rightarrow48$ & $+1.62$\\
26B & enabled  & $5\rightarrow2$  & $8\rightarrow8$   & $-3.67$\\
26B & disabled & $4\rightarrow4$  & $42\rightarrow42$ & $+0.04$\\
31B & enabled  & $13\rightarrow3$ & $11\rightarrow13$ & $+5.04$\\
31B & disabled & $0\rightarrow0$  & $46\rightarrow47$ & $-0.17$\\
\bottomrule
\end{tabular}
\caption{Held-out outcomes, each row covering 6 prompts $\times$ 8 unseen seeds.
``Loops'' and ``natural EOS'' are counts out of 48. $\Delta$ unique is the
selected-minus-baseline mean number of unique parsed list items.}
\label{tab:heldout_outcomes}
\end{table}

The U.S.-states control is preserved almost perfectly: 127/128 generations
contain all 50 states and aggregate coverage is 6398/6400 items. The important
negative result is E2B with thinking enabled. Its selected edit produces fewer
unique items on the longer chemical-elements and Pok\'emon prompts and reaches
natural EOS in 33/48 rather than 44/48 runs. Pooled across both modes, the E2B
change is $-6.84$ unique items (95\% bootstrap CI [$-12.31$,$-2.28$]) and
$-9.4$\,pp natural EOS ([$-16.7$,$-2.1$]). The held-out result therefore
supports loop suppression, not an unconditional quality improvement.

\subsection{Matched Random-Intervention Controls}

For each model, we freeze five matched random interventions before evaluation.
Dense controls use the same layers, intervention directions, and unit counts as
the selected edit; 26B controls mask three routed experts with the same
allocation of two experts at L21 and one at L22. Selected units are excluded from every draw. Each
random intervention is evaluated on the same 48 held-out generations per
thinking mode. Table~\ref{tab:random_controls} reports the resulting empirical
distribution; the five draws are controls, not enough samples for a calibrated
randomization test.

\begin{table}[htbp]
\centering\small
\setlength{\tabcolsep}{5pt}
\begin{tabular}{@{}llrrrrr@{}}
\toprule
\textbf{model} & \textbf{thinking} & \shortstack{\textbf{selected}\\\textbf{loops}} &
\shortstack{\textbf{random}\\\textbf{loop rate}} & \shortstack{\textbf{draws}\\\textbf{$\leq$S}} &
\shortstack{\textbf{natural}\\\textbf{EOS S/R}} & \shortstack{\textbf{unique}\\\textbf{items S/R}}\\
\midrule
E2B & enabled  & 1/48 & .013 [.000, .042] & 4/5 & .688/.879 & 39.5/47.9\\
E2B & disabled & 0/48 & .083 [.062, .104] & 0/5 & 1.000/.963 & 47.6/48.1\\
E4B & enabled  & 2/48 & .133 [.083, .167] & 0/5 & .812/.825 & 46.9/43.6\\
E4B & disabled & 0/48 & .171 [.146, .208] & 0/5 & 1.000/.988 & 52.2/50.6\\
26B & enabled  & 2/48 & .050 [.021, .083] & 2/5 & .167/.175 & 71.0/73.8\\
26B & disabled & 4/48 & .067 [.042, .104] & 4/5 & .875/.867 & 55.5/55.6\\
31B & enabled  & 3/48 & .175 [.104, .229] & 0/5 & .271/.225 & 31.1/30.0\\
31B & disabled & 0/48 & .004 [.000, .021] & 4/5 & .979/.929 & 55.5/55.4\\
\bottomrule
\end{tabular}
\caption{Selected edits versus five matched random interventions. Random entries
are mean loop rate [minimum, maximum]. ``$\leq$S'' counts random draws whose loop
rate is no greater than the selected rate. EOS and unique-item columns report
selected/random means.}
\label{tab:random_controls}
\end{table}

The selected edit beats every matched draw on loop incidence for E4B in both
modes, E2B without thinking, and 31B with thinking. The 31B no-thinking setting is
at a floor for selected and random edits. In contrast, selected 26B and
thinking-enabled E2B lie inside their matched-random ranges; those settings do not
support a claim that the attributed units are uniquely selective. We therefore
treat attribution as candidate generation except where the selected edit
separates from these controls.

\subsection{Blinded Detector Audit}

We drew a stratified audit sample of 81 outputs from the four primary failure
probes (Firefly, \textit{The Wire}, constellations, and Generation-I
Pok\'emon), their held-out paraphrases, and six enumeration controls (U.S.
states, U.S. presidents, chemical elements, EU member states, noble gases, and
MCU films). The sample spans baseline and edited models, the initial and
long-budget evaluations, detected tight and soft loops, and detector-negative
outputs that either ended naturally or exhausted the token budget. Reviewers
saw only the generated text and operational definitions, without detector
outcomes. For this validity audit, a human repetition-positive judgment is the
union of tight loops, soft list collapse, and a broader long-block
phrase-repetition label. Initial binary judgments agreed on 69/81 outputs
(85.2\%); a
detector-blind third judgment resolved the remaining 12 disagreements. The
resulting majority reference contains 44 repetition-positive and 37 negative
outputs.

As a separate model-assisted sensitivity check, a fresh instance of GPT-5.6
Sol (\texttt{gpt-5.6-sol-medium}) independently reviewed the same blinded texts. It was not given the
paper-detector, Antidoom, human, or prior model annotations. Its binary labels
agree with the final human reference on 75/81 outputs (92.6\%; Cohen's
$\kappa=0.852$). This model-based audit is a sensitivity check, not a
substitute for the primary human reference.

\begin{table}[htbp]
\centering\small
\begin{tabular}{llcc}
\toprule
\textbf{evaluated method} & \textbf{reference labels} &
\textbf{TP/FP/FN/TN} & \textbf{P/R/F1} \\
\midrule
Ours & Human majority & 36/5/8/32 & .878/.818/.847 \\
Antidoom & Human majority & 26/1/18/36 & .963/.591/.732 \\
Ours & GPT-5.6 Sol & 37/4/3/37 & .902/.925/.914 \\
\bottomrule
\end{tabular}
\caption{Blinded repetition-audit results on the same stratified 81-output
sample. The majority human reference is primary; GPT-5.6 is a separate
model-assisted sensitivity check.}
\label{tab:detector_audit}
\end{table}

Table~\ref{tab:detector_audit} reports the resulting comparisons. For our
detector against the human reference, four of the five false positives
arise when structural verification lines normalize to empty content and are
mistaken for list collapse; the fifth is a long separator-character run that
the human reference classifies as other degeneration. All eight false negatives
contain long repeated enumeration, restart, or verification blocks that exceed
the 40-token period bound or vary across repetitions without producing four
consecutive identical normalized entries.

Antidoom's sole false positive is the same separator run; its 18 false
negatives comprise 12 normalized list collapses and six long-range
block-repetition cases. Thus Antidoom is more conservative on this audit, while
our detector captures more of the human-labeled cases. Because the sample is
detector-stratified, these are audit-sample metrics, not population prevalence
estimates or proof of detector superiority.

\subsection{Untouched Capability Benchmarks}

We remove MMLU from the reported development benchmark figure because its
task-level short answer limit truncates Gemma's prose-first answers. As an
independent check, we evaluate HellaSwag, WinoGrande, and BoolQ without thinking
on 500 examples each; none was used to select an intervention.
Table~\ref{tab:independent_benchmarks} gives selected-minus-baseline changes.
The intervals conservatively combine the two reported standard errors as if
independent; paired intervals would be narrower. Every interval includes zero,
so these runs show no measurable shift but have limited precision.

\begin{table}[htbp]
\centering\small
\begin{tabular}{llrrr}
\toprule
\textbf{model} & \textbf{benchmark} & \textbf{baseline} & \textbf{selected} &
\textbf{$\Delta$ pp [95\% CI]}\\
\midrule
E2B & HellaSwag  & .500 & .494 & $-0.6$ [$-6.8$, $+5.6$]\\
E2B & WinoGrande & .568 & .572 & $+0.4$ [$-5.7$, $+6.5$]\\
E2B & BoolQ      & .808 & .774 & $-3.4$ [$-8.4$, $+1.6$]\\
E4B & HellaSwag  & .676 & .682 & $+0.6$ [$-5.2$, $+6.4$]\\
E4B & WinoGrande & .690 & .696 & $+0.6$ [$-5.1$, $+6.3$]\\
E4B & BoolQ      & .802 & .842 & $+4.0$ [$-0.7$, $+8.7$]\\
26B & HellaSwag  & .714 & .722 & $+0.8$ [$-4.8$, $+6.4$]\\
26B & WinoGrande & .816 & .816 & $0.0$ [$-4.8$, $+4.8$]\\
26B & BoolQ      & .882 & .896 & $+1.4$ [$-2.5$, $+5.3$]\\
31B & HellaSwag  & .756 & .758 & $+0.2$ [$-5.1$, $+5.5$]\\
31B & WinoGrande & .916 & .908 & $-0.8$ [$-4.3$, $+2.7$]\\
31B & BoolQ      & .898 & .894 & $-0.4$ [$-4.2$, $+3.4$]\\
\bottomrule
\end{tabular}
\caption{Untouched no-thinking capability benchmarks (500 examples per model--benchmark comparison).
Scores are proportions; deltas and conservative 95\% CIs are percentage points.}
\label{tab:independent_benchmarks}
\end{table}

\FloatBarrier
\section{Localization and Attribution Details}
\label{appendix:localization}

\begin{figure}[htbp]
  \centering
  \includegraphics[width=0.95\textwidth]{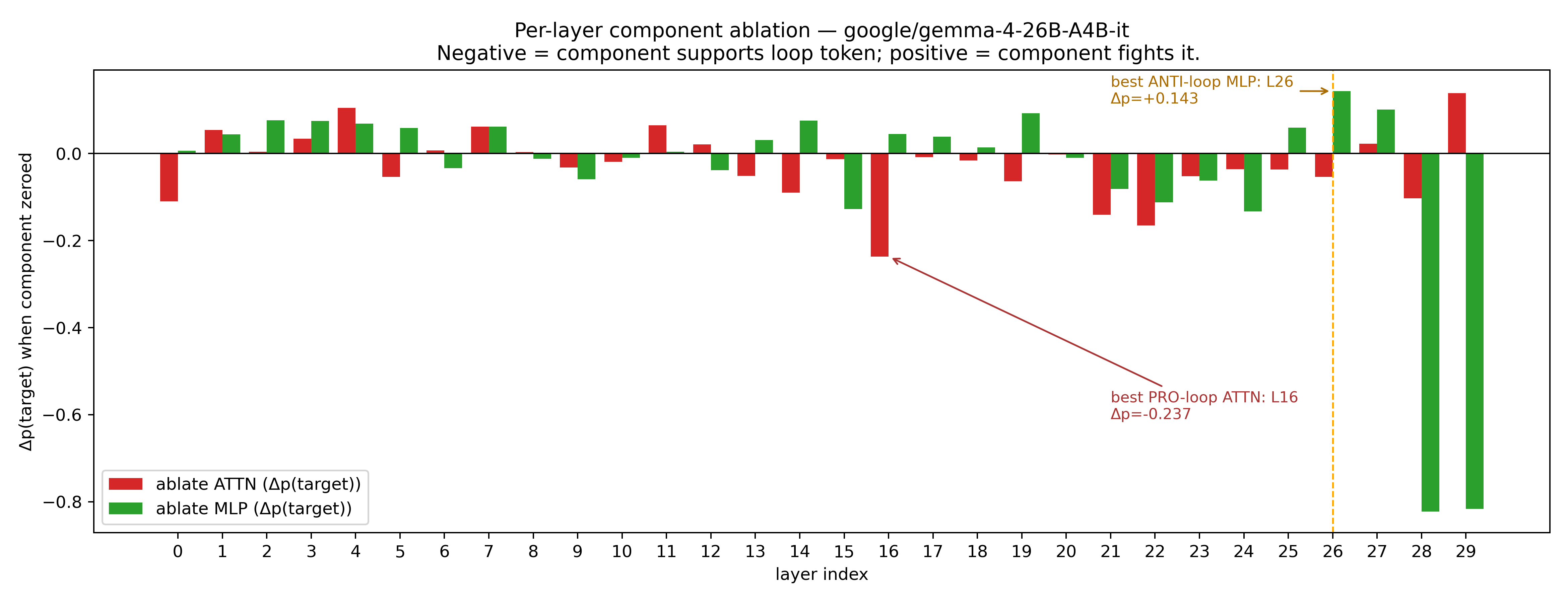}
  \caption{Per-layer component ablation for the 26B model. Each bar pair shows the change in $p(\text{loop-token} \mid \text{context})$ when attention (red, left) or MLP (green, right) is zero-ablated at that layer; negative values mean the component supports the loop token, positive values mean it suppresses it. The strongest pro-loop attention signal is at L16 ($\Delta p = -0.237$); the strongest anti-loop MLP signal is at L26 ($\Delta p = +0.143$, dashed line), with additional large pro-loop MLP signal at L28--29 that reflects the final write-out layers rather than loop-specific structure. The expert-attribution analysis (Figure~\ref{moe_loops_fig}) then drills into the L21--22 region to identify the specific expert positions that drive the loop. E2B, E4B, and 31B are analyzed with the same ablation procedure; E2B illustrates that the strongest single-token write signal need not coincide with the most effective intervention layer.}
  \label{fig:ablation_26b}
\end{figure}

Figure~\ref{fig:ablation_26b} shows the full per-layer ablation sweep for 26B, and
Table~\ref{antiloop_tab} summarizes the candidate and intervention layers across all
four models. These diagnostics guide candidate selection, while the final edits are
chosen by full-generation behavioral sweeps.

\subsection{Dense-MLP Attribution Score}

For the E2B, E4B, and 31B models, we score every MLP neuron in the candidate
layer~$\ell$. Gemma 4 text MLPs use a GeGLU~\citep{shazeer2020glu,dauphin2017language}
gating structure:
\[
\mathbf{h}=W_{\!\downarrow}\!\Bigl(\phi(W_g\mathbf{x})\odot W_u\mathbf{x}\Bigr),
\]
where $\phi$ is GELU~\citep{hendrycks2016gaussian} (tanh approximation), and
$W_{\!\downarrow}$, $W_g$, and $W_u$ denote \texttt{down\_proj},
\texttt{gate\_proj}, and \texttt{up\_proj}. We define neuron~$n$ at layer~$\ell$
as index position $n$ in the intermediate dimension. All interventions target
column~$n$ of $W_{\!\downarrow}^{(\ell)}$ directly, modifying the neuron's write
direction while leaving its read directions and scalar activation unchanged.
At each position $p$, we record the post-gate activation
\[
a_n^{(\ell)}(p)=\phi\!\left(W_g^{(\ell)}[n]\cdot\mathbf{x}_p\right)
\cdot\left(W_u^{(\ell)}[n]\cdot\mathbf{x}_p\right),
\]
and the gradient of the target loop-token log-probability:
\[
g_n^{(\ell)}(p)=
\frac{\partial\log p_\theta(t\mid x_{<p})}{\partial a_n^{(\ell)}(p)}.
\]
The loop-attribution score is
\[
\Delta_n^{(\ell)}=-\sum_p a_n^{(\ell)}(p)\,g_n^{(\ell)}(p)
\approx
\log p_{\theta^{(\ell,n,0)}}(t)-\log p_\theta(t),
\]
where $\theta^{(\ell,n,0)}$ denotes the model with neuron $n$'s
\texttt{down\_proj} column zeroed. This gradient-times-activation approximation
scores all neurons in one forward and backward pass. A positive score identifies
an anti-loop neuron (zeroing would raise loop-token probability), while a negative
score identifies a pro-loop neuron. We confirm top-$K$ candidates by exact
zero-ablation and full-generation evaluation on the prompt--seed grid.

\begin{figure}[htbp]
  \centering
  \includegraphics[width=0.95\textwidth]{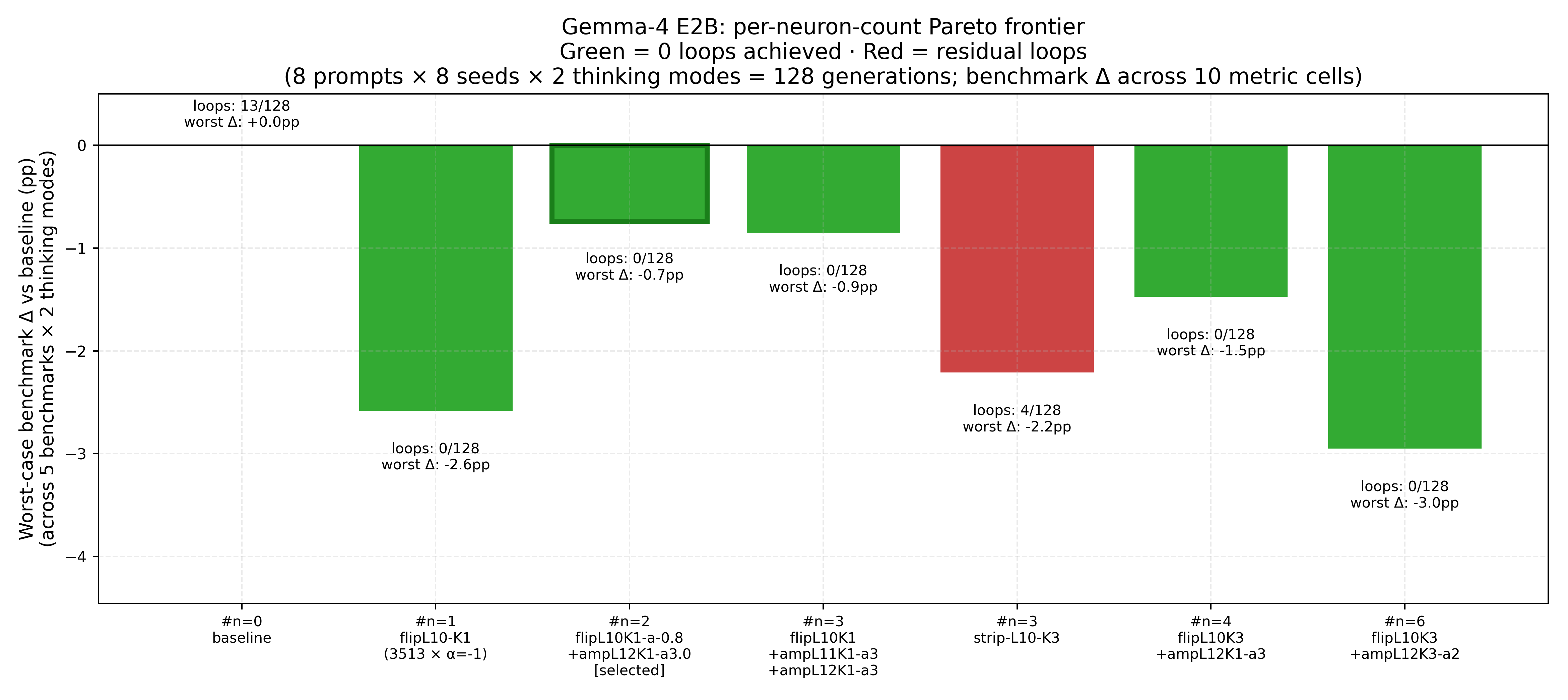}
  \caption{E2B per-neuron-count Pareto frontier. Each bar is a weight-edit
  variant plotted at its worst-case regression across five valid benchmarks
  under both thinking modes. Annotations report loop counts on the original prompt-and-seed set;
  the selected two-neuron edit (bold border) reaches 0/128 loops with a
  $-0.74$\,pp worst-case benchmark change.}
  \label{fig:e2b_pareto}
\end{figure}

\begin{table}[htbp]
\centering\small
\renewcommand{\arraystretch}{1.3}
\begin{tabularx}{\textwidth}{@{}p{2.1cm}p{1.8cm}p{4.1cm}X p{2.2cm}@{}}
\toprule
\textbf{model} & \textbf{layers} & \textbf{ablation / intervention layer} & \textbf{which} & \textbf{depth} \\
\midrule
\textbf{31B} & 60 (dense) & \textbf{L36} & MLP \texttt{down\_proj} & 60\% \\
\textbf{E4B} & 42 (dense) & \textbf{L18} & MLP \texttt{down\_proj} & 43\% \\
\textbf{E2B} & 35 (dense) & \textbf{L13--L15} / \textbf{L10, L12} & MLP \texttt{down\_proj} (ablation: L13--L15; intervention: L10, L12) & 37--43\% / 29--34\% \\
\textbf{26B} & 30 (MoE) & \textbf{L21--L22} & routed experts & 70--73\% \\
\bottomrule
\end{tabularx}
\caption{Candidate layers identified by per-layer ablation and final intervention layers for each model. For 31B and E4B the ablation peak coincides with the intervention layer. For E2B, ablation identifies L13--L15 as the site of strongest loop-token signal, while the selected intervention acts at L10/L12 (see main paper Sec.~4). For 26B, the ablation signal peaks across L21--L26, and the selected expert masks are applied at L21--L22.}
\label{antiloop_tab}
\end{table}

\subsection{Attribution Stability Across Looping Captures}

We apply this stability analysis to two representative dense models with
single-layer neuron edits: E4B, which has the strongest matched-control
evidence, and 31B, the largest dense model. E2B's selected intervention spans
two layers, while 26B edits routed experts, so neither fits the same
single-layer dense-neuron ranking protocol. We froze the protocol before
computing these results. Candidate
generations were baseline, thinking-enabled detector positives from seeds
0--7. We then recaptured each loop through HF Transformers, which is required
for gradient attribution, and computed both the per-layer ablation ranking and
the full neuron ranking at the selected intervention layer. The analysis uses
six reproducible captures for each model. All six planned E4B captures were
usable. For 31B, six of eight were usable: two \texttt{constellations}
generations that looped in the vLLM selection run did not loop when recaptured
through the HF attribution path and were excluded before any attribution score
was computed.

The rankings vary substantially across captures. The strongest pro-loop
ablation layer did not match the selected intervention layer in any of the six
captures for either model. Within the same prompt, full neuron-rank Spearman
correlation and top-100 overlap average 0.056/0.097 for E4B and 0.122/0.205 for
31B. Cross-prompt values are 0.007/0.081 for E4B and 0.138/0.198 for 31B.
Accordingly, attribution provides a useful candidate generator, but the
full-generation sweeps, held-out evaluations, and matched controls, rather than
ranking stability, support the selected edits.

\FloatBarrier
\section{Intervention-Search Details}
\label{appendix:intervention_search}

\subsection{31B Probe-Combination Strategies}

Table~\ref{31b_comb_tab} compares selection rules for the L36 stripping edit. The
result shows that prompt-specific drivers from two attractors are more useful than
the literal intersection of neurons ranked highly on both prompts.

\begin{table}[htbp]
\centering\small
\renewcommand{\arraystretch}{1.3}
\begin{tabularx}{\textwidth}{@{}p{3.2cm}p{4.1cm}p{1.8cm}X@{}}
\toprule
\textbf{strategy} & \textbf{selection criterion} & \textbf{best result} & \textbf{notes} \\
\midrule
\texttt{wireonly-K1000}
  & top-$K$ from the \texttt{wire\_episodes} probe alone
  & 4\,/\,64
  & Reduces failures from 13/64 to 4/64; the \texttt{wire\_episodes} probe elicits a sharper loop attractor than the \texttt{firefly\_list} probe, making its top-ranked neurons more discriminative. \\[4pt]
\texttt{wirelong-}\newline\texttt{intersect-}\newline\texttt{K10/16/33}
  & neurons in the top-100 (or top-200) of \textit{both} the \texttt{firefly\_list} and \texttt{wire\_episodes} probes
  & 9--11\,/\,64
  & Intersection underperforms: retaining only neurons that rank highly on both probes discards the prompt-specific drivers that sustain each attractor independently. \\[4pt]
\texttt{union200-K367}
  & union of top-200 \texttt{firefly\_list} $\cup$ top-200 \texttt{wire\_episodes} (367 unique neurons)
  & 6\,/\,64
  & Marginally effective but inferior to the additive sum strategy; includes many low-attribution neurons from each probe. \\[4pt]
\texttt{sum-K1000}
  & rank by \texttt{score\_firefly + score\_wire}, take top-$K$
  & 1\,/\,64
  & Additive combination concentrates on neurons that score highly on both probes without requiring them to rank in the top of each individually. \\[4pt]
\texttt{maxpos-K1000}
  & rank by the maximum of the two prompt scores, take top-$K$
  & 1\,/\,64
  & Equivalent result to \texttt{sum} for this neuron set. \\[4pt]
\hcode{strip-ff-}\newline\hcode{K1000-wonly-}\newline\hcode{K100}
  & strip top-1000 from \texttt{firefly\_list} $\cup$ a \textit{separate} top-100 from \texttt{wire\_episodes}
  & \textbf{0\,/\,64\,$\star$}
  & \textbf{Selected variant}: allows each probe to contribute its highest-ranked neurons independently, without forcing both probe signals through a shared ranking. The 1,100 ranked selections contain 38 overlaps, so 1,062 unique neurons are stripped. \\
\bottomrule
\end{tabularx}
\caption{Probe-combination strategies for the 31B L36 stripping edit. Each row is a neuron selection strategy evaluated on the $8{\times}8$ grid (64 generations, thinking enabled); \textbf{best result} is the lowest loop count achieved across the tested values of $K$ (number of neurons stripped; the best-performing $K$ for each strategy is encoded in the variant name); $\star$ marks the selected variant (which retains 1/64 residual in thinking-disabled mode).}
\label{31b_comb_tab}
\end{table}

The intersection numbers indicate that neurons that are loop-driving on
\textit{every} trigger are not necessarily the right target. Different prompts can
enter different attractor basins in the same layer. For example,
\texttt{firefly\_list} often collapses toward ``\textit{The Message},'' while
\texttt{wire\_episodes} often collapses toward ``\textit{The Detail}.'' The neurons
supporting those attractors overlap only partially. The selected intervention
therefore strips both sets, with $K$ tuned per probe: 1000 neurons from
\texttt{firefly\_list} and 100 from \texttt{wire\_episodes}.

\FloatBarrier
\subsection{26B Routed-Expert Attribution and Masking}

\begin{figure}[htbp]
  \centering
  \includegraphics[width=0.95\textwidth]{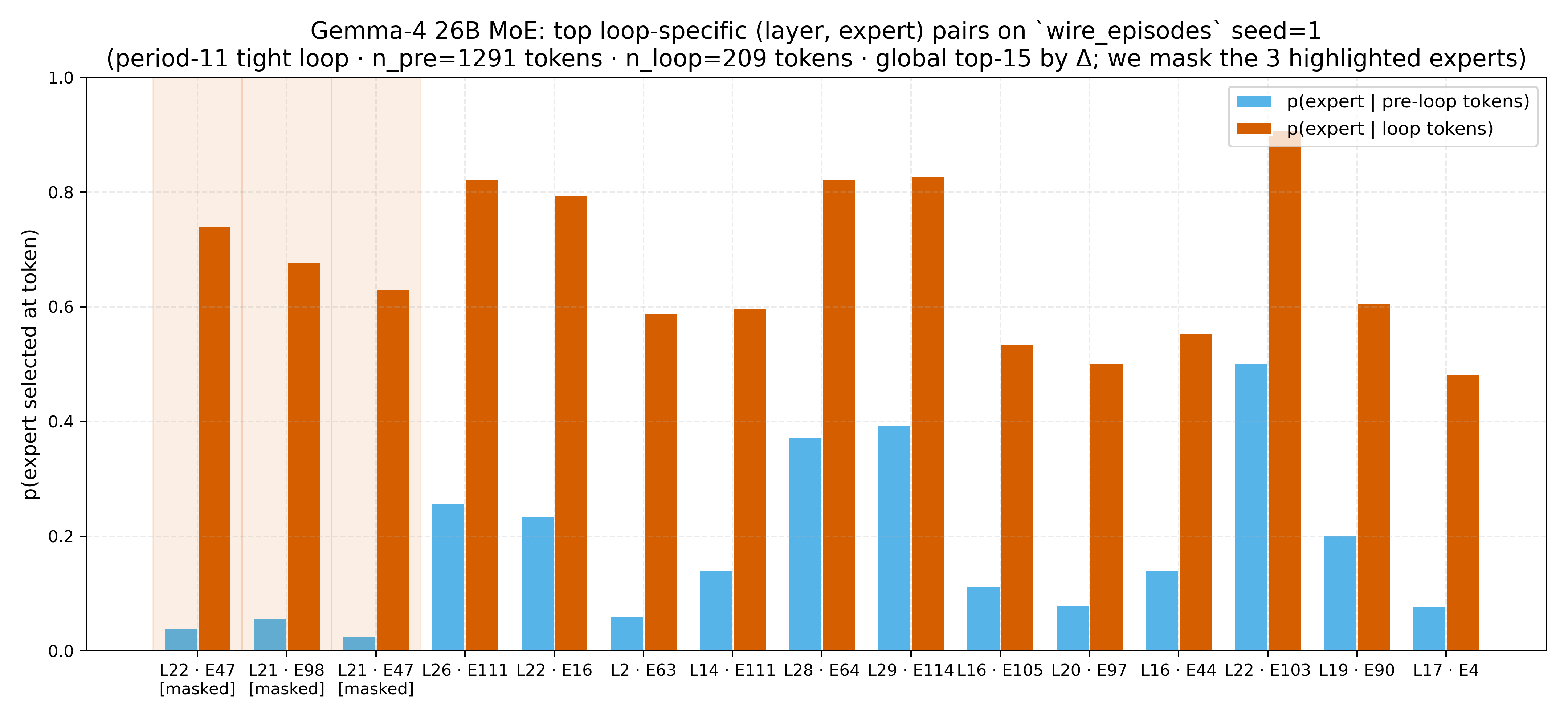}
  \caption{Top-15 (layer, expert) pairs ranked by $\Delta = p(\text{expert} \mid \text{loop}) - p(\text{expert} \mid \text{pre-loop})$, computed on a single 26B rollout of \texttt{wire\_episodes} seed~1 (period-11 tight loop; 1{,}291 pre-loop and 209 loop tokens). The three \emph{[masked]} experts (L22:E47, L21:E98, L21:E47) jump from under 6\% selection pre-loop to over 60\% inside the loop; masking these three yields the \texttt{v2\_top3} intervention (Figure~\ref{moe_mask_fig}).}
  \label{moe_loops_fig}
\end{figure}

\begin{figure}[htbp]
  \centering
  \includegraphics[width=0.95\textwidth]{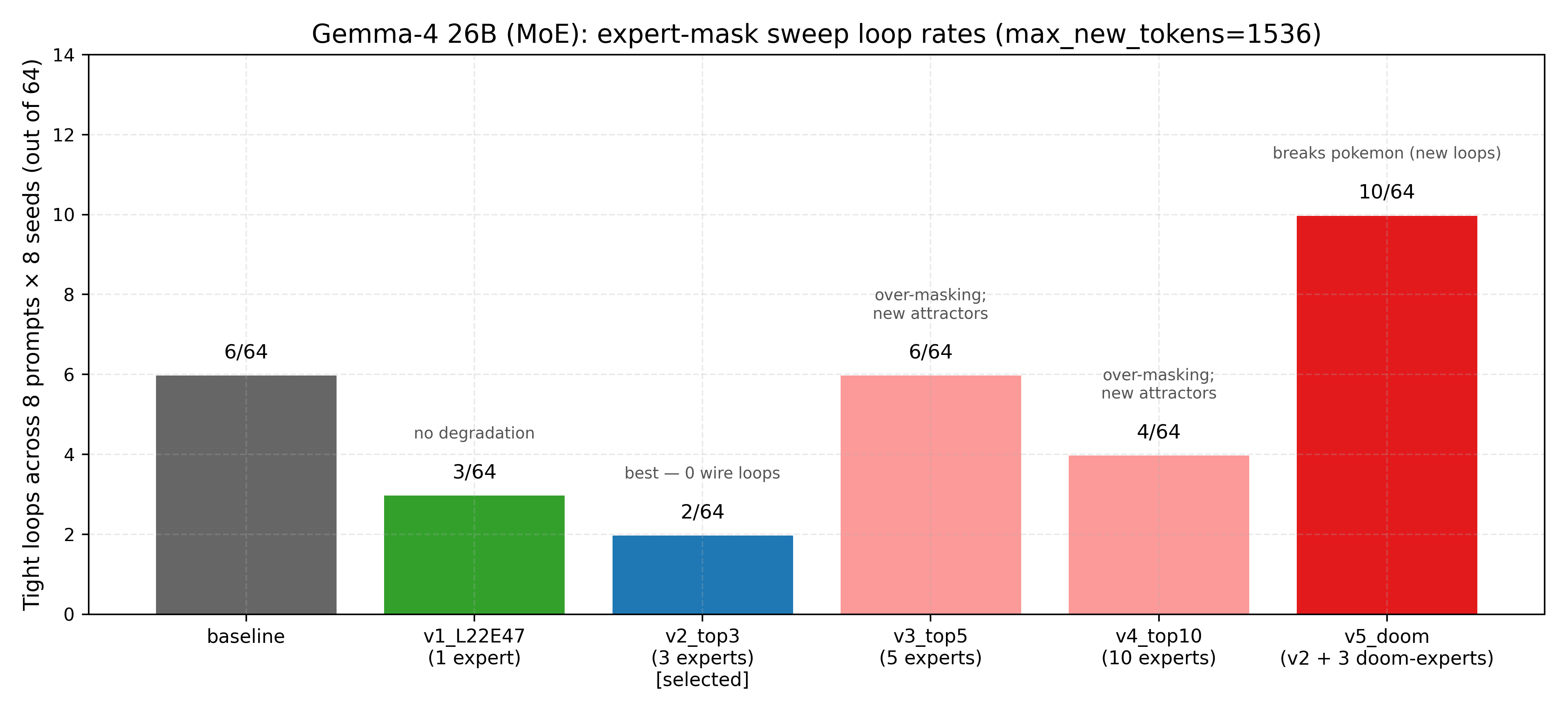}
  \caption{Tight-loop counts ($8\text{ prompts} \times 8\text{ seeds} = 64$ generations, \texttt{max\_new\_tokens=1536}, thinking enabled) under five expert-mask interventions. \texttt{v2\_top3} (L21:E47, L21:E98, L22:E47) is the selected variant, reducing tight loops from 6/64 to 2/64. Masking more experts (\texttt{v3\_top5}, \texttt{v4\_top10}) does not improve further as the model re-routes to alternate attractors. \texttt{v5\_doom} targets experts active during long-budget rumination but proves to suppress general-purpose reasoning, introducing new loops on \texttt{pokemon\_gen1} (5/8 seeds).}
  \label{moe_mask_fig}
\end{figure}

The 26B model uses routed experts for most FFN computation, so we compare expert
selection frequencies before and during a captured loop. Figure~\ref{moe_loops_fig}
shows the resulting ranking; Figure~\ref{moe_mask_fig} evaluates progressively larger
masks. Both figures derive candidate units from one captured
\texttt{wire\_episodes} rollout and validate interventions on the full prompt--seed
grid.

For each layer $L$ and routed expert $E$, the loop-specificity score is
\[
\Delta_E^{(L)}
=
\frac{1}{|T_{\mathrm{loop}}|}\sum_{p\in T_{\mathrm{loop}}}
\mathbf{1}\!\left[E \in \mathrm{TopK}\!\left(g_L(x_{<p})\right)\right]
-
\frac{1}{|T_{\mathrm{pre}}|}\sum_{p\in T_{\mathrm{pre}}}
\mathbf{1}\!\left[E \in \mathrm{TopK}\!\left(g_L(x_{<p})\right)\right].
\]
Here, $g_L(x_{<p})$ denotes the router logits at layer $L$ for position $p$, and
\(\mathrm{TopK}\) is the selected expert set. A large positive score means the
expert is selected much more often inside the loop than before it begins.

The expert mask is unconditional: it is active for every token on every prompt.
A conditional mask would require custom inference code and an online detector,
whereas our goal is a static drop-in weight modification. This constraint makes
the sweep delicate: the objective is not to suppress every expert involved in
looping, but to find the smallest loop-specific set that can be permanently zeroed
without degrading non-loop generations.

\FloatBarrier
\subsection{General-Purpose Benchmark Details}

\begin{figure}[htbp]
  \centering
  \includegraphics[width=0.95\textwidth]{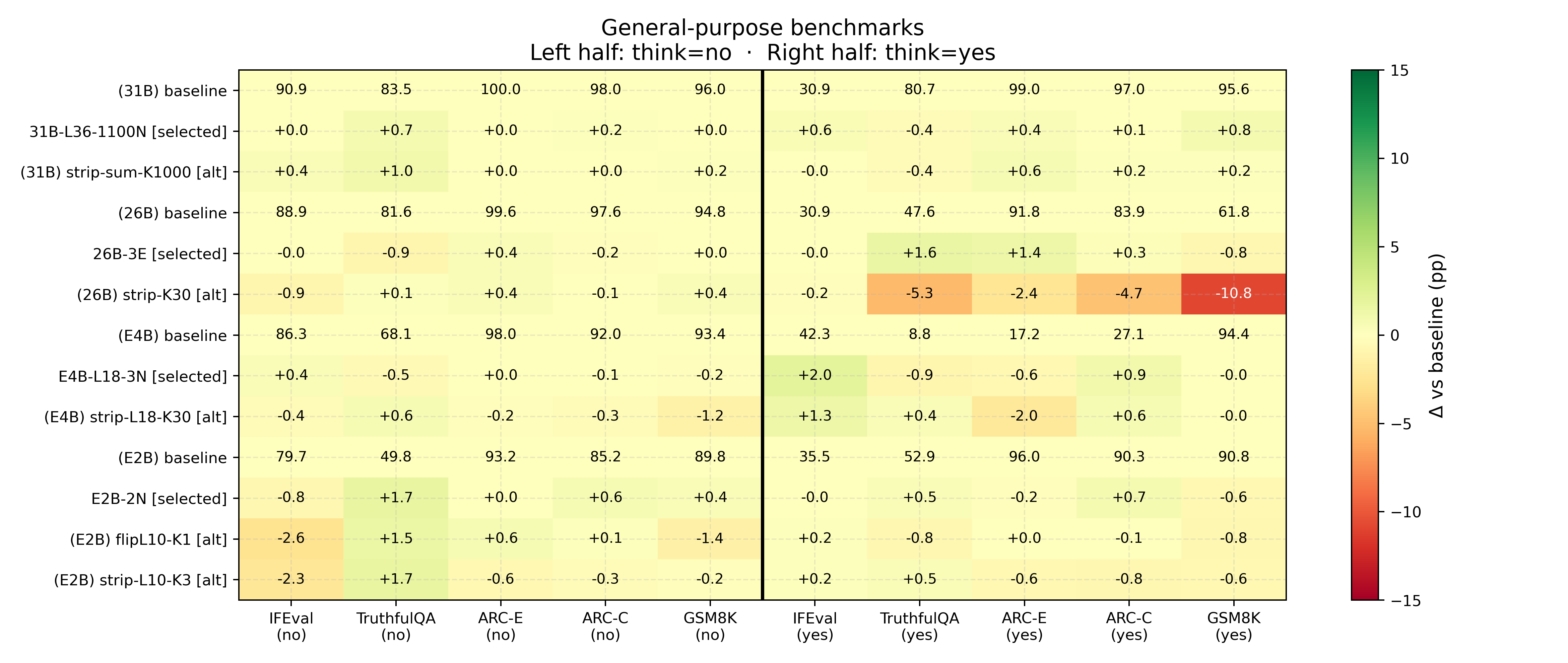}
  \caption{Benchmark performance deltas relative to the unedited baseline, across five valid general-purpose benchmarks and two thinking modes. Green\,=\,improvement over baseline; red\,=\,regression. The four selected variants (one per model) are highlighted. Five alternative model variants are included for comparison. MMLU is excluded because its task-level short answer cap truncates these models' prose-first responses.}
  \label{benchmarks_fig}
\end{figure}

Figure~\ref{benchmarks_fig} reports every benchmark result used to compare the
selected edits with the unedited models, together with the alternative variants
considered during intervention search. The compact table in the main paper reports
only the selected variants' meaningful ranges; the full figure is retained here for
auditability.

\FloatBarrier
\section{Full Per-Generation Enumeration Results}
\label{appendix:full_enum}

Table~\ref{tab:full_enum} shows the loop verdict for every combination of model, variant, thinking mode, prompt, and seed in the $8\text{-prompt} \times 8\text{-seed} \times 2\text{-mode}$ evaluation. The same loop detector is applied uniformly to all 128 generations per model variant. Glyph meanings: \textbf{T} = tight loop; \textbf{L} = soft loop (list-collapse); $\bullet$ = no loop.

\providecommand{\tightL}{{\color{red!70!black}\textbf{T}}}
\providecommand{\listL}{{\color{orange!85!black}\textbf{L}}}
\providecommand{\nolp}{{\color{gray}\(\bullet\)}}
\providecommand{\pending}{{\color{gray}\textendash}}
\providecommand{\hcode}[1]{\texttt{#1}}

{\small\setlength{\tabcolsep}{1pt}\renewcommand{\arraystretch}{1.05}
\begin{longtable}{@{}p{1.4cm}>{\raggedright\arraybackslash}p{2.8cm}*{8}{c}*{3}{c}@{}}
\toprule
\textbf{setting} & \textbf{prompt} &
  \textbf{s0} & \textbf{s1} & \textbf{s2} & \textbf{s3} &
  \textbf{s4} & \textbf{s5} & \textbf{s6} & \textbf{s7} &
  \textbf{T} & \textbf{soft} & \textbf{n/N} \\
\midrule
\endfirsthead
\multicolumn{13}{l}{\textit{(continued)}}\\
\toprule
\textbf{setting} & \textbf{prompt} &
  \textbf{s0} & \textbf{s1} & \textbf{s2} & \textbf{s3} &
  \textbf{s4} & \textbf{s5} & \textbf{s6} & \textbf{s7} &
  \textbf{T} & \textbf{soft} & \textbf{n/N} \\
\midrule
\endhead
\endfoot
\multicolumn{13}{@{}l}{\textbf{\texttt{gemma-4-E2B-it}}} \\
\midrule
\multirow{8}{=}{\textbf{Baseline}, \textsf{think=no}}
  & \texttt{constellations} & \listL & \listL & \listL & \listL & \listL & \listL & \listL & \nolp & 0 & 7 & 7/8 \\
  & \texttt{firefly} & \nolp & \nolp & \nolp & \nolp & \nolp & \nolp & \nolp & \nolp & 0 & 0 & 0/8 \\
  & \texttt{wire} & \nolp & \nolp & \nolp & \nolp & \nolp & \nolp & \nolp & \nolp & 0 & 0 & 0/8 \\
  & \texttt{pokemon} & \nolp & \nolp & \nolp & \nolp & \nolp & \nolp & \nolp & \nolp & 0 & 0 & 0/8 \\
  & \texttt{us\_pres} & \nolp & \nolp & \nolp & \nolp & \nolp & \nolp & \nolp & \nolp & 0 & 0 & 0/8 \\
  & \texttt{mcu} & \nolp & \nolp & \nolp & \nolp & \nolp & \nolp & \nolp & \nolp & 0 & 0 & 0/8 \\
  & \texttt{eu\_states} & \nolp & \nolp & \nolp & \nolp & \nolp & \nolp & \nolp & \nolp & 0 & 0 & 0/8 \\
  & \texttt{noble\_gas} & \nolp & \nolp & \nolp & \nolp & \nolp & \nolp & \nolp & \nolp & 0 & 0 & 0/8 \\
  & \multicolumn{1}{r}{\textbf{Total}} & \multicolumn{8}{c}{} & 0 & 7 & 7/64 \\[1pt]
\midrule
\multirow{8}{=}{\textbf{Baseline}, \textsf{think=yes}}
  & \texttt{constellations} & \listL & \listL & \listL & \nolp & \listL & \nolp & \listL & \nolp & 0 & 5 & 5/8 \\
  & \texttt{firefly} & \nolp & \listL & \nolp & \nolp & \nolp & \nolp & \nolp & \nolp & 0 & 1 & 1/8 \\
  & \texttt{wire} & \nolp & \nolp & \nolp & \nolp & \nolp & \nolp & \nolp & \nolp & 0 & 0 & 0/8 \\
  & \texttt{pokemon} & \nolp & \nolp & \nolp & \nolp & \nolp & \nolp & \nolp & \nolp & 0 & 0 & 0/8 \\
  & \texttt{us\_pres} & \nolp & \nolp & \nolp & \nolp & \nolp & \nolp & \nolp & \nolp & 0 & 0 & 0/8 \\
  & \texttt{mcu} & \nolp & \nolp & \nolp & \nolp & \nolp & \nolp & \nolp & \nolp & 0 & 0 & 0/8 \\
  & \texttt{eu\_states} & \nolp & \nolp & \nolp & \nolp & \nolp & \nolp & \nolp & \nolp & 0 & 0 & 0/8 \\
  & \texttt{noble\_gas} & \nolp & \nolp & \nolp & \nolp & \nolp & \nolp & \nolp & \nolp & 0 & 0 & 0/8 \\
  & \multicolumn{1}{r}{\textbf{Total}} & \multicolumn{8}{c}{} & 0 & 6 & 6/64 \\[1pt]
\midrule
\multirow{8}{=}{\textbf{Selected}, \textsf{think=no}}
  & \texttt{constellations} & \nolp & \nolp & \nolp & \nolp & \nolp & \nolp & \nolp & \nolp & 0 & 0 & 0/8 \\
  & \texttt{firefly} & \nolp & \nolp & \nolp & \nolp & \nolp & \nolp & \nolp & \nolp & 0 & 0 & 0/8 \\
  & \texttt{wire} & \nolp & \nolp & \nolp & \nolp & \nolp & \nolp & \nolp & \nolp & 0 & 0 & 0/8 \\
  & \texttt{pokemon} & \nolp & \nolp & \nolp & \nolp & \nolp & \nolp & \nolp & \nolp & 0 & 0 & 0/8 \\
  & \texttt{us\_pres} & \nolp & \nolp & \nolp & \nolp & \nolp & \nolp & \nolp & \nolp & 0 & 0 & 0/8 \\
  & \texttt{mcu} & \nolp & \nolp & \nolp & \nolp & \nolp & \nolp & \nolp & \nolp & 0 & 0 & 0/8 \\
  & \texttt{eu\_states} & \nolp & \nolp & \nolp & \nolp & \nolp & \nolp & \nolp & \nolp & 0 & 0 & 0/8 \\
  & \texttt{noble\_gas} & \nolp & \nolp & \nolp & \nolp & \nolp & \nolp & \nolp & \nolp & 0 & 0 & 0/8 \\
  & \multicolumn{1}{r}{\textbf{Total}} & \multicolumn{8}{c}{} & 0 & 0 & 0/64 \\[1pt]
\midrule
\multirow{8}{=}{\textbf{Selected}, \textsf{think=yes}}
  & \texttt{constellations} & \nolp & \nolp & \nolp & \nolp & \nolp & \nolp & \nolp & \nolp & 0 & 0 & 0/8 \\
  & \texttt{firefly} & \nolp & \nolp & \nolp & \nolp & \nolp & \nolp & \nolp & \nolp & 0 & 0 & 0/8 \\
  & \texttt{wire} & \nolp & \nolp & \nolp & \nolp & \nolp & \nolp & \nolp & \nolp & 0 & 0 & 0/8 \\
  & \texttt{pokemon} & \nolp & \nolp & \nolp & \nolp & \nolp & \nolp & \nolp & \nolp & 0 & 0 & 0/8 \\
  & \texttt{us\_pres} & \nolp & \nolp & \nolp & \nolp & \nolp & \nolp & \nolp & \nolp & 0 & 0 & 0/8 \\
  & \texttt{mcu} & \nolp & \nolp & \nolp & \nolp & \nolp & \nolp & \nolp & \nolp & 0 & 0 & 0/8 \\
  & \texttt{eu\_states} & \nolp & \nolp & \nolp & \nolp & \nolp & \nolp & \nolp & \nolp & 0 & 0 & 0/8 \\
  & \texttt{noble\_gas} & \nolp & \nolp & \nolp & \nolp & \nolp & \nolp & \nolp & \nolp & 0 & 0 & 0/8 \\
  & \multicolumn{1}{r}{\textbf{Total}} & \multicolumn{8}{c}{} & 0 & 0 & 0/64 \\[1pt]
\midrule
\multicolumn{13}{@{}l}{\textbf{\texttt{gemma-4-E4B-it}}} \\
\midrule
\multirow{8}{=}{\textbf{Baseline}, \textsf{think=no}}
  & \texttt{constellations} & \nolp & \nolp & \listL & \nolp & \nolp & \nolp & \nolp & \nolp & 0 & 1 & 1/8 \\
  & \texttt{firefly} & \listL & \listL & \listL & \listL & \nolp & \listL & \listL & \listL & 0 & 7 & 7/8 \\
  & \texttt{wire} & \nolp & \nolp & \nolp & \nolp & \nolp & \nolp & \nolp & \nolp & 0 & 0 & 0/8 \\
  & \texttt{pokemon} & \nolp & \nolp & \nolp & \nolp & \nolp & \nolp & \nolp & \nolp & 0 & 0 & 0/8 \\
  & \texttt{us\_pres} & \nolp & \nolp & \nolp & \nolp & \nolp & \nolp & \nolp & \nolp & 0 & 0 & 0/8 \\
  & \texttt{mcu} & \nolp & \nolp & \nolp & \nolp & \nolp & \nolp & \nolp & \nolp & 0 & 0 & 0/8 \\
  & \texttt{eu\_states} & \nolp & \nolp & \nolp & \nolp & \nolp & \nolp & \nolp & \nolp & 0 & 0 & 0/8 \\
  & \texttt{noble\_gas} & \nolp & \nolp & \nolp & \nolp & \nolp & \nolp & \nolp & \nolp & 0 & 0 & 0/8 \\
  & \multicolumn{1}{r}{\textbf{Total}} & \multicolumn{8}{c}{} & 0 & 8 & 8/64 \\[1pt]
\midrule
\multirow{8}{=}{\textbf{Baseline}, \textsf{think=yes}}
  & \texttt{constellations} & \nolp & \nolp & \nolp & \nolp & \nolp & \nolp & \tightL & \tightL & 2 & 0 & 2/8 \\
  & \texttt{firefly} & \nolp & \nolp & \nolp & \nolp & \listL & \listL & \listL & \listL & 0 & 4 & 4/8 \\
  & \texttt{wire} & \nolp & \nolp & \nolp & \nolp & \nolp & \nolp & \nolp & \nolp & 0 & 0 & 0/8 \\
  & \texttt{pokemon} & \nolp & \nolp & \nolp & \nolp & \nolp & \nolp & \nolp & \nolp & 0 & 0 & 0/8 \\
  & \texttt{us\_pres} & \nolp & \nolp & \nolp & \nolp & \nolp & \nolp & \nolp & \nolp & 0 & 0 & 0/8 \\
  & \texttt{mcu} & \nolp & \nolp & \nolp & \nolp & \nolp & \nolp & \nolp & \nolp & 0 & 0 & 0/8 \\
  & \texttt{eu\_states} & \nolp & \nolp & \nolp & \nolp & \nolp & \nolp & \nolp & \nolp & 0 & 0 & 0/8 \\
  & \texttt{noble\_gas} & \nolp & \nolp & \nolp & \nolp & \nolp & \nolp & \nolp & \nolp & 0 & 0 & 0/8 \\
  & \multicolumn{1}{r}{\textbf{Total}} & \multicolumn{8}{c}{} & 2 & 4 & 6/64 \\[1pt]
\midrule
\multirow{8}{=}{\textbf{Selected}, \textsf{think=no}}
  & \texttt{constellations} & \nolp & \nolp & \listL & \nolp & \nolp & \nolp & \nolp & \nolp & 0 & 1 & 1/8 \\
  & \texttt{firefly} & \nolp & \nolp & \nolp & \nolp & \nolp & \nolp & \listL & \nolp & 0 & 1 & 1/8 \\
  & \texttt{wire} & \nolp & \nolp & \nolp & \nolp & \nolp & \nolp & \nolp & \nolp & 0 & 0 & 0/8 \\
  & \texttt{pokemon} & \nolp & \nolp & \nolp & \nolp & \nolp & \nolp & \nolp & \nolp & 0 & 0 & 0/8 \\
  & \texttt{us\_pres} & \nolp & \nolp & \nolp & \nolp & \nolp & \nolp & \nolp & \nolp & 0 & 0 & 0/8 \\
  & \texttt{mcu} & \nolp & \nolp & \nolp & \nolp & \nolp & \nolp & \nolp & \nolp & 0 & 0 & 0/8 \\
  & \texttt{eu\_states} & \nolp & \nolp & \nolp & \nolp & \nolp & \nolp & \nolp & \nolp & 0 & 0 & 0/8 \\
  & \texttt{noble\_gas} & \nolp & \nolp & \nolp & \nolp & \nolp & \nolp & \nolp & \nolp & 0 & 0 & 0/8 \\
  & \multicolumn{1}{r}{\textbf{Total}} & \multicolumn{8}{c}{} & 0 & 2 & 2/64 \\[1pt]
\midrule
\multirow{8}{=}{\textbf{Selected}, \textsf{think=yes}}
  & \texttt{constellations} & \nolp & \nolp & \nolp & \nolp & \nolp & \nolp & \nolp & \nolp & 0 & 0 & 0/8 \\
  & \texttt{firefly} & \nolp & \nolp & \nolp & \nolp & \nolp & \nolp & \nolp & \nolp & 0 & 0 & 0/8 \\
  & \texttt{wire} & \nolp & \nolp & \nolp & \nolp & \nolp & \nolp & \nolp & \nolp & 0 & 0 & 0/8 \\
  & \texttt{pokemon} & \nolp & \nolp & \nolp & \nolp & \nolp & \nolp & \nolp & \nolp & 0 & 0 & 0/8 \\
  & \texttt{us\_pres} & \nolp & \nolp & \nolp & \nolp & \nolp & \nolp & \nolp & \nolp & 0 & 0 & 0/8 \\
  & \texttt{mcu} & \nolp & \nolp & \nolp & \nolp & \nolp & \nolp & \nolp & \nolp & 0 & 0 & 0/8 \\
  & \texttt{eu\_states} & \nolp & \nolp & \nolp & \nolp & \nolp & \nolp & \nolp & \nolp & 0 & 0 & 0/8 \\
  & \texttt{noble\_gas} & \nolp & \nolp & \nolp & \nolp & \nolp & \nolp & \nolp & \nolp & 0 & 0 & 0/8 \\
  & \multicolumn{1}{r}{\textbf{Total}} & \multicolumn{8}{c}{} & 0 & 0 & 0/64 \\[1pt]
\midrule
\multicolumn{13}{@{}l}{\textbf{\texttt{gemma-4-31B-it}}} \\
\midrule
\multirow{8}{=}{\textbf{Baseline}, \textsf{think=no}}
  & \texttt{constellations} & \nolp & \nolp & \nolp & \nolp & \nolp & \nolp & \nolp & \nolp & 0 & 0 & 0/8 \\
  & \texttt{firefly} & \nolp & \nolp & \nolp & \nolp & \tightL & \nolp & \nolp & \nolp & 1 & 0 & 1/8 \\
  & \texttt{wire} & \nolp & \nolp & \nolp & \nolp & \nolp & \nolp & \nolp & \nolp & 0 & 0 & 0/8 \\
  & \texttt{pokemon} & \nolp & \nolp & \nolp & \nolp & \nolp & \nolp & \nolp & \nolp & 0 & 0 & 0/8 \\
  & \texttt{us\_pres} & \nolp & \nolp & \nolp & \nolp & \nolp & \nolp & \nolp & \nolp & 0 & 0 & 0/8 \\
  & \texttt{mcu} & \nolp & \nolp & \nolp & \nolp & \nolp & \nolp & \nolp & \nolp & 0 & 0 & 0/8 \\
  & \texttt{eu\_states} & \nolp & \nolp & \nolp & \nolp & \nolp & \nolp & \nolp & \nolp & 0 & 0 & 0/8 \\
  & \texttt{noble\_gas} & \nolp & \nolp & \nolp & \nolp & \nolp & \nolp & \nolp & \nolp & 0 & 0 & 0/8 \\
  & \multicolumn{1}{r}{\textbf{Total}} & \multicolumn{8}{c}{} & 1 & 0 & 1/64 \\[1pt]
\midrule
\multirow{8}{=}{\textbf{Baseline}, \textsf{think=yes}}
  & \texttt{constellations} & \nolp & \listL & \listL & \listL & \nolp & \listL & \nolp & \listL & 0 & 5 & 5/8 \\
  & \texttt{firefly} & \nolp & \nolp & \nolp & \nolp & \tightL & \nolp & \tightL & \tightL & 3 & 0 & 3/8 \\
  & \texttt{wire} & \nolp & \tightL & \nolp & \tightL & \listL & \tightL & \nolp & \tightL & 4 & 1 & 5/8 \\
  & \texttt{pokemon} & \nolp & \nolp & \nolp & \nolp & \nolp & \nolp & \nolp & \nolp & 0 & 0 & 0/8 \\
  & \texttt{us\_pres} & \nolp & \nolp & \nolp & \nolp & \nolp & \nolp & \nolp & \nolp & 0 & 0 & 0/8 \\
  & \texttt{mcu} & \nolp & \nolp & \nolp & \nolp & \nolp & \nolp & \nolp & \nolp & 0 & 0 & 0/8 \\
  & \texttt{eu\_states} & \nolp & \nolp & \nolp & \nolp & \nolp & \nolp & \nolp & \nolp & 0 & 0 & 0/8 \\
  & \texttt{noble\_gas} & \nolp & \nolp & \nolp & \nolp & \nolp & \nolp & \nolp & \nolp & 0 & 0 & 0/8 \\
  & \multicolumn{1}{r}{\textbf{Total}} & \multicolumn{8}{c}{} & 7 & 6 & 13/64 \\[1pt]
\midrule
\multirow{8}{=}{\textbf{Selected}, \textsf{think=no}}
  & \texttt{constellations} & \nolp & \nolp & \nolp & \nolp & \nolp & \nolp & \nolp & \nolp & 0 & 0 & 0/8 \\
  & \texttt{firefly} & \nolp & \nolp & \nolp & \tightL & \nolp & \nolp & \nolp & \nolp & 1 & 0 & 1/8 \\
  & \texttt{wire} & \nolp & \nolp & \nolp & \nolp & \nolp & \nolp & \nolp & \nolp & 0 & 0 & 0/8 \\
  & \texttt{pokemon} & \nolp & \nolp & \nolp & \nolp & \nolp & \nolp & \nolp & \nolp & 0 & 0 & 0/8 \\
  & \texttt{us\_pres} & \nolp & \nolp & \nolp & \nolp & \nolp & \nolp & \nolp & \nolp & 0 & 0 & 0/8 \\
  & \texttt{mcu} & \nolp & \nolp & \nolp & \nolp & \nolp & \nolp & \nolp & \nolp & 0 & 0 & 0/8 \\
  & \texttt{eu\_states} & \nolp & \nolp & \nolp & \nolp & \nolp & \nolp & \nolp & \nolp & 0 & 0 & 0/8 \\
  & \texttt{noble\_gas} & \nolp & \nolp & \nolp & \nolp & \nolp & \nolp & \nolp & \nolp & 0 & 0 & 0/8 \\
  & \multicolumn{1}{r}{\textbf{Total}} & \multicolumn{8}{c}{} & 1 & 0 & 1/64 \\[1pt]
\midrule
\multirow{8}{=}{\textbf{Selected}, \textsf{think=yes}}
  & \texttt{constellations} & \nolp & \nolp & \nolp & \nolp & \nolp & \nolp & \nolp & \nolp & 0 & 0 & 0/8 \\
  & \texttt{firefly} & \nolp & \nolp & \nolp & \nolp & \nolp & \nolp & \nolp & \nolp & 0 & 0 & 0/8 \\
  & \texttt{wire} & \nolp & \nolp & \nolp & \nolp & \nolp & \nolp & \nolp & \nolp & 0 & 0 & 0/8 \\
  & \texttt{pokemon} & \nolp & \nolp & \nolp & \nolp & \nolp & \nolp & \nolp & \nolp & 0 & 0 & 0/8 \\
  & \texttt{us\_pres} & \nolp & \nolp & \nolp & \nolp & \nolp & \nolp & \nolp & \nolp & 0 & 0 & 0/8 \\
  & \texttt{mcu} & \nolp & \nolp & \nolp & \nolp & \nolp & \nolp & \nolp & \nolp & 0 & 0 & 0/8 \\
  & \texttt{eu\_states} & \nolp & \nolp & \nolp & \nolp & \nolp & \nolp & \nolp & \nolp & 0 & 0 & 0/8 \\
  & \texttt{noble\_gas} & \nolp & \nolp & \nolp & \nolp & \nolp & \nolp & \nolp & \nolp & 0 & 0 & 0/8 \\
  & \multicolumn{1}{r}{\textbf{Total}} & \multicolumn{8}{c}{} & 0 & 0 & 0/64 \\[1pt]
\midrule
\multicolumn{13}{@{}l}{\textbf{\texttt{gemma-4-26B-A4B-it}}} \\
\midrule
\multirow{8}{=}{\textbf{Baseline}, \textsf{think=no}}
  & \texttt{constellations} & \nolp & \nolp & \nolp & \nolp & \nolp & \nolp & \nolp & \nolp & 0 & 0 & 0/8 \\
  & \texttt{firefly} & \nolp & \tightL & \nolp & \nolp & \nolp & \nolp & \nolp & \nolp & 1 & 0 & 1/8 \\
  & \texttt{wire} & \nolp & \nolp & \nolp & \tightL & \tightL & \nolp & \tightL & \nolp & 3 & 0 & 3/8 \\
  & \texttt{pokemon} & \nolp & \nolp & \nolp & \nolp & \nolp & \nolp & \nolp & \nolp & 0 & 0 & 0/8 \\
  & \texttt{us\_pres} & \nolp & \nolp & \nolp & \nolp & \nolp & \nolp & \nolp & \nolp & 0 & 0 & 0/8 \\
  & \texttt{mcu} & \nolp & \nolp & \nolp & \nolp & \nolp & \nolp & \nolp & \nolp & 0 & 0 & 0/8 \\
  & \texttt{eu\_states} & \nolp & \nolp & \nolp & \nolp & \nolp & \nolp & \nolp & \nolp & 0 & 0 & 0/8 \\
  & \texttt{noble\_gas} & \nolp & \nolp & \nolp & \nolp & \nolp & \nolp & \nolp & \nolp & 0 & 0 & 0/8 \\
  & \multicolumn{1}{r}{\textbf{Total}} & \multicolumn{8}{c}{} & 4 & 0 & 4/64 \\[1pt]
\midrule
\multirow{8}{=}{\textbf{Baseline}, \textsf{think=yes}}
  & \texttt{constellations} & \nolp & \nolp & \nolp & \nolp & \nolp & \nolp & \nolp & \nolp & 0 & 0 & 0/8 \\
  & \texttt{firefly} & \nolp & \nolp & \nolp & \tightL & \nolp & \nolp & \nolp & \nolp & 1 & 0 & 1/8 \\
  & \texttt{wire} & \nolp & \tightL & \tightL & \tightL & \tightL & \tightL & \nolp & \nolp & 5 & 0 & 5/8 \\
  & \texttt{pokemon} & \nolp & \nolp & \nolp & \nolp & \nolp & \nolp & \nolp & \nolp & 0 & 0 & 0/8 \\
  & \texttt{us\_pres} & \nolp & \nolp & \nolp & \nolp & \nolp & \nolp & \nolp & \nolp & 0 & 0 & 0/8 \\
  & \texttt{mcu} & \nolp & \nolp & \nolp & \nolp & \nolp & \nolp & \nolp & \nolp & 0 & 0 & 0/8 \\
  & \texttt{eu\_states} & \nolp & \nolp & \nolp & \nolp & \nolp & \nolp & \nolp & \nolp & 0 & 0 & 0/8 \\
  & \texttt{noble\_gas} & \nolp & \nolp & \nolp & \nolp & \nolp & \nolp & \nolp & \nolp & 0 & 0 & 0/8 \\
  & \multicolumn{1}{r}{\textbf{Total}} & \multicolumn{8}{c}{} & 6 & 0 & 6/64 \\[1pt]
\midrule
\multirow{8}{=}{\textbf{Selected}, \textsf{think=no}}
  & \texttt{constellations} & \nolp & \nolp & \nolp & \nolp & \nolp & \nolp & \nolp & \nolp & 0 & 0 & 0/8 \\
  & \texttt{firefly} & \nolp & \nolp & \nolp & \nolp & \nolp & \nolp & \nolp & \nolp & 0 & 0 & 0/8 \\
  & \texttt{wire} & \nolp & \nolp & \nolp & \nolp & \nolp & \tightL & \nolp & \nolp & 1 & 0 & 1/8 \\
  & \texttt{pokemon} & \nolp & \nolp & \nolp & \nolp & \nolp & \nolp & \nolp & \nolp & 0 & 0 & 0/8 \\
  & \texttt{us\_pres} & \nolp & \nolp & \nolp & \nolp & \nolp & \nolp & \nolp & \nolp & 0 & 0 & 0/8 \\
  & \texttt{mcu} & \nolp & \nolp & \nolp & \nolp & \nolp & \nolp & \nolp & \nolp & 0 & 0 & 0/8 \\
  & \texttt{eu\_states} & \nolp & \nolp & \nolp & \nolp & \nolp & \nolp & \nolp & \nolp & 0 & 0 & 0/8 \\
  & \texttt{noble\_gas} & \nolp & \nolp & \nolp & \nolp & \nolp & \nolp & \nolp & \nolp & 0 & 0 & 0/8 \\
  & \multicolumn{1}{r}{\textbf{Total}} & \multicolumn{8}{c}{} & 1 & 0 & 1/64 \\[1pt]
\midrule
\multirow{8}{=}{\textbf{Selected}, \textsf{think=yes}}
  & \texttt{constellations} & \nolp & \nolp & \nolp & \nolp & \nolp & \nolp & \nolp & \nolp & 0 & 0 & 0/8 \\
  & \texttt{firefly} & \nolp & \listL & \nolp & \tightL & \nolp & \nolp & \nolp & \nolp & 1 & 1 & 2/8 \\
  & \texttt{wire} & \nolp & \nolp & \nolp & \nolp & \nolp & \nolp & \nolp & \nolp & 0 & 0 & 0/8 \\
  & \texttt{pokemon} & \nolp & \nolp & \nolp & \nolp & \nolp & \nolp & \nolp & \nolp & 0 & 0 & 0/8 \\
  & \texttt{us\_pres} & \nolp & \nolp & \nolp & \nolp & \nolp & \nolp & \nolp & \nolp & 0 & 0 & 0/8 \\
  & \texttt{mcu} & \nolp & \nolp & \nolp & \nolp & \nolp & \nolp & \nolp & \nolp & 0 & 0 & 0/8 \\
  & \texttt{eu\_states} & \nolp & \nolp & \nolp & \nolp & \nolp & \nolp & \nolp & \nolp & 0 & 0 & 0/8 \\
  & \texttt{noble\_gas} & \nolp & \nolp & \nolp & \nolp & \nolp & \nolp & \nolp & \nolp & 0 & 0 & 0/8 \\
  & \multicolumn{1}{r}{\textbf{Total}} & \multicolumn{8}{c}{} & 1 & 1 & 2/64 \\[1pt]
\bottomrule
\caption{Loop verdict for every (model, variant, thinking mode, prompt, seed) cell in the
8-prompt $\times$ 8-seed $\times$ 2-mode evaluation grid, for both the unpatched
baseline and the selected variant of each model.
Each cell reports the loop classification for a single generation:
\tightL{}\,=\,tight token-period loop,
\listL{}\,=\,numbered-list collapse,
\nolp{}\,=\,no loop.
The \textbf{T} column counts tight loops; \textbf{soft} counts list-collapse loops;
\textbf{n/N} is the total over all eight seeds. Exact selected-variant identifiers
appear in Table~\ref{variant_names_tab}.}
\label{tab:full_enum} \\
\end{longtable}}

\paragraph{Note on the auxiliary rendered-text check (\textbf{P}).}
Our loop detector includes a narrow rendered-text check that fires when an exact
8--120-character unit repeats contiguously at least four times at the output
tail without forming a 5--40-token period. It is an implementation-level
backstop rather than a third paper phenotype. No output in the reported initial
or long-budget evaluations is classified by this check (\textbf{P} does not
appear in the table above), but it fires on a small number of residual outputs
from non-selected variants. A representative example occurs in one E2B variant
on the \texttt{constellations} probe:

\begin{quote}\small\ttfamily
  {\ldots}\\
  69.\ Volans Australis Septentrionalis Meridionalis Minor Minor Minor {\ldots} Minor\\
  70.\ Volans Australis Septentrionalis Meridionalis Minor Minor Minor {\ldots} Minor Major\\
  71.\ Volans Australis Septentrionalis Meridionalis Minor Minor Minor {\ldots} Minor\\
  72.\ Volans Australis Septentrionalis Meridionalis Minor Minor Minor {\ldots} Minor Major
\end{quote}

The inner phrase repeats across lines, but the leading line number and trailing \texttt{Minor}/\texttt{Major} token vary each cycle, so the tight detector (which requires exact token-ID periodicity) does not fire; the phrase detector catches the repeated substring instead.

\section{Long-Budget Doom-Looping Outcomes}
\label{appendix:doom_outcomes}

Table~\ref{tab:doom_outcomes} reports outcome shares for the two doom-prone models (26B and 31B) on the three doom-prone prompts (\texttt{wire\_episodes}, \texttt{firefly\_list}, \texttt{constellations}), pooled over both 4k and 8k budgets with \texttt{think=yes}. Each row covers $8\,\text{seeds} \times 3\,\text{prompts} \times 2\,\text{budgets} = 48$ generations. Columns: \textbf{tight} = tight loop running to budget; \textbf{soft} = soft loop running to budget; \textbf{endless} = budget exhausted without a verbatim lock and without a natural end-of-sequence token (the model was still self-correcting when the budget ran out); \textbf{doom} = tight\,+\,soft\,+\,endless; \textbf{loop+nat} = detector-flagged as a loop but terminated naturally (manual verification); \textbf{nat-EOS} = natural end-of-sequence completion. E2B and E4B are excluded; both reach $\geq$\,46/48 natural-EOS in every long-budget setting.

\begin{table}[htbp]
  \centering\small
  \setlength{\tabcolsep}{5pt}
  \begin{tabular}{@{}llrrrrrrr@{}}
  \toprule
  \textbf{model} & \textbf{variant} & \textbf{rep\_pen} & \textbf{tight} & \textbf{soft} & \textbf{endless} & \textbf{doom} & \textbf{loop+nat} & \textbf{nat-EOS} \\
  \midrule
  26B & baseline & 1.00 & 17 & 0 & 17 & \textbf{34} & 0 & 14 \\
  26B & baseline & 1.15 & 12 & 1 &  6 & \textbf{19} & 0 & 29 \\
  26B & selected & 1.00 &  6 & 4 & 29 & \textbf{39} & 0 &  9 \\
  26B & selected & 1.15 & 14 & 0 &  2 & \textbf{16} & 0 & 32 \\
  \midrule
  31B & baseline & 1.00 & 24 & 2 &  6 & \textbf{32} & 0 & 16 \\
  31B & baseline & 1.15 &  4 & 9 & 17 & \textbf{30} & 0 & 18 \\
  31B & selected & 1.00 & 10 & 2 & 26 & \textbf{38} & 0 & 10 \\
  31B & selected & 1.15 &  2 & 5 & 19 & \textbf{26} & 2 & 20 \\
  \bottomrule
  \end{tabular}
  \caption{Doom-looping outcome shares for 26B and 31B on the three doom-prone prompts, pooled over 4k and 8k budgets (n\,=\,48 per row, think=yes). See main paper Sec.~7.1 for discussion. Note: 8 entries in the 31B baseline \texttt{rep\_pen=1.0} row are detector-flagged soft loops on \texttt{constellations} that are in fact clean completions; the model writes an alphabetical-coverage verification step in its scratchpad (``W:(none), X:(none), Y:(none), Z:(none)'') that the soft-loop detector flags, but it still emits a complete 88-line list and terminates naturally; one further such case appears in the 31B selected \texttt{rep\_pen=1.15} row. All are counted in nat-EOS rather than in the loop columns.}
  \label{tab:doom_outcomes}
\end{table}

\begin{figure}[htbp]
  \centering
  \includegraphics[width=0.95\textwidth]{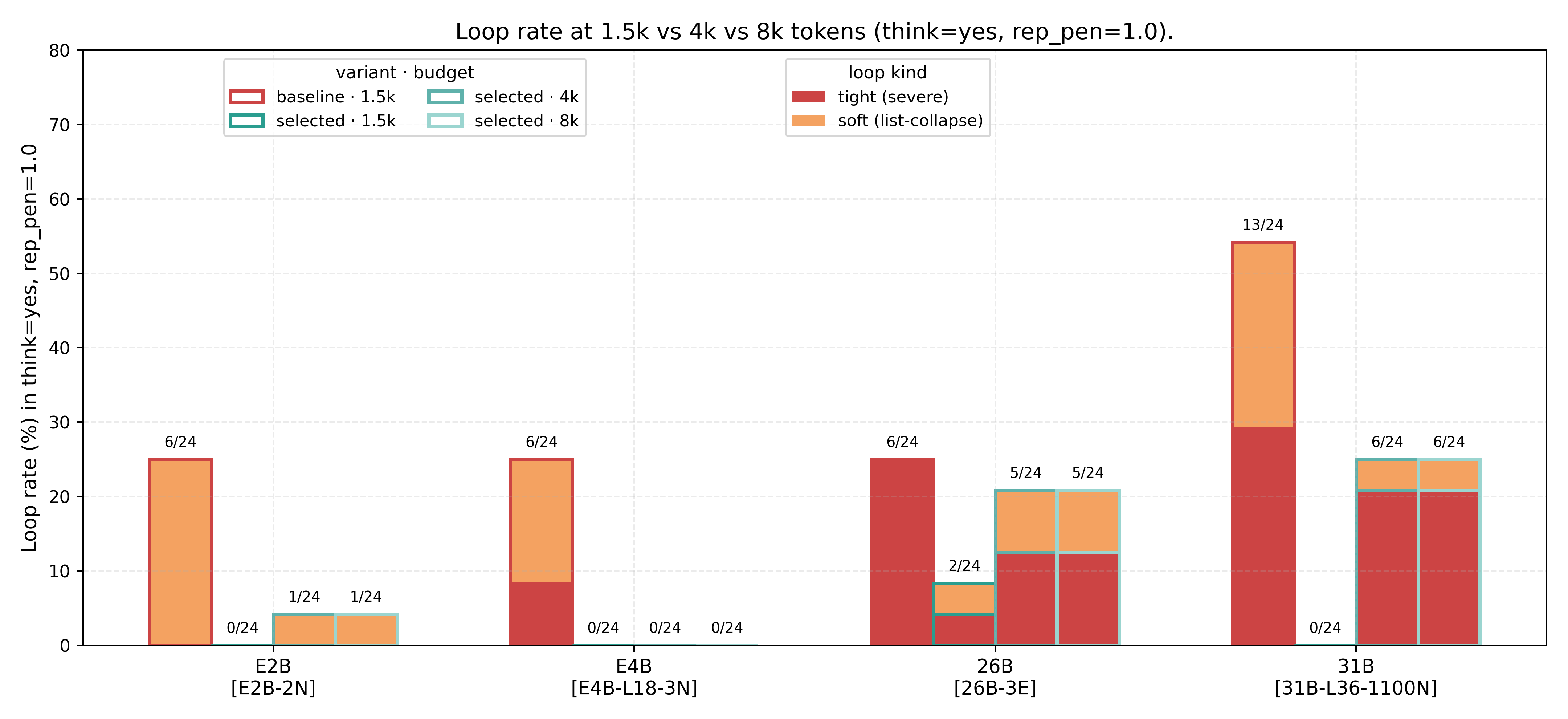}
  \caption{Selected-variant loop rates at 1.5k, 4k, and 8k token budgets
  (\texttt{think=yes}, \texttt{rep\_pen=1.0}), pooled over 3 doom-prone prompts
  $\times$ 8 seeds per setting ($n{=}24$, all models). E2B and E4B remain clean at
  long budgets ($\leq 1/24$ residual, all soft). 26B and 31B exhibit substantial
  doom looping that does not worsen between 4k and 8k but does not resolve either.}
  \label{fig:long_budget_doom}
\end{figure}

\begin{figure}[htbp]
  \centering
  \includegraphics[width=0.95\textwidth]{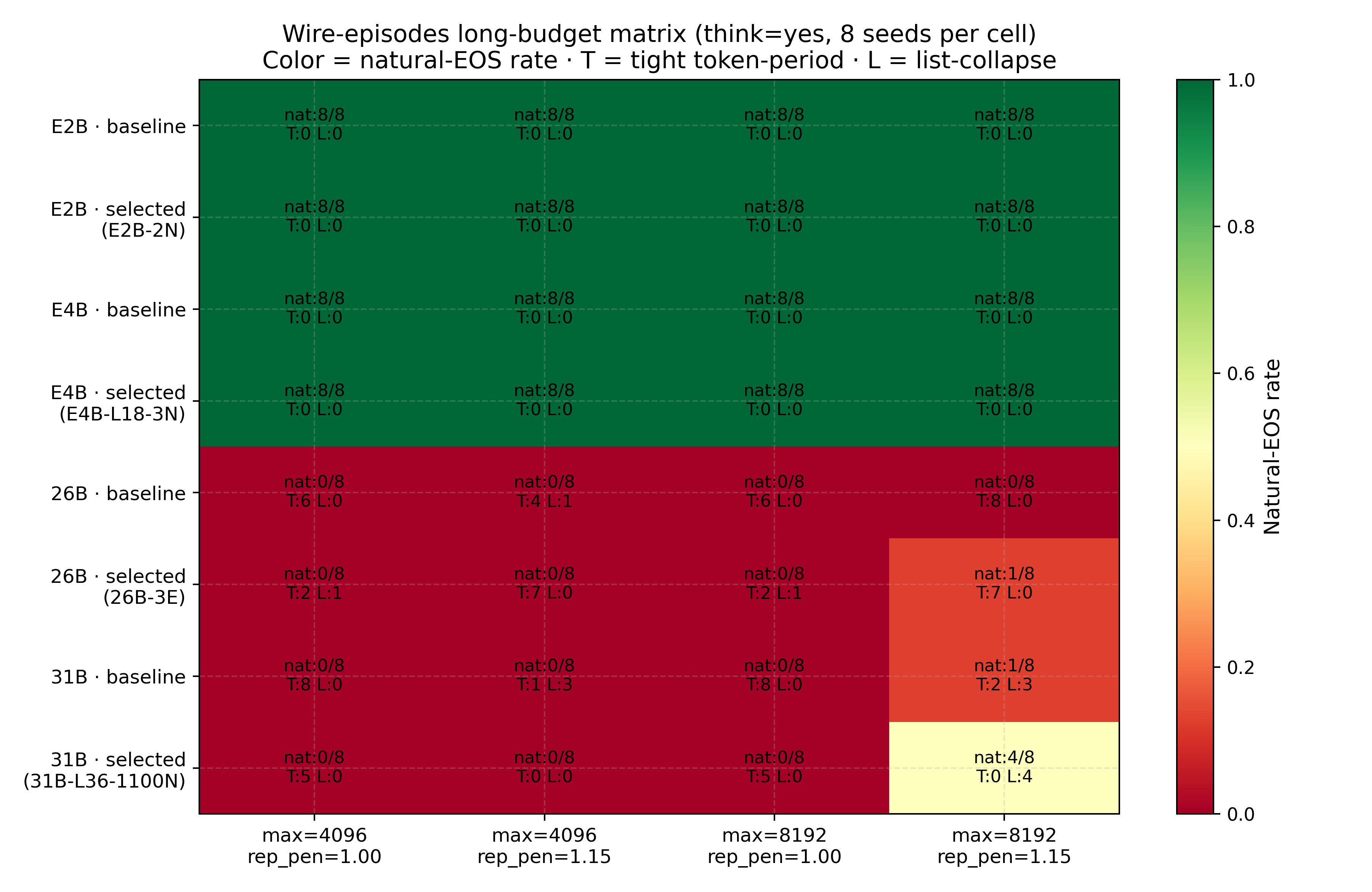}
  \caption{Natural-EOS (End-of-Sequence) rate (cell colour) on the
  \texttt{wire\_episodes} prompt across all four models, both variants (baseline and
  selected intervention), and four long-budget cells (2 token budgets $\times$ 2
  repetition penalties), \texttt{think=yes}, 8 seeds per cell. Cell annotations
  show \textbf{nat}:N/8 (natural-EOS count), \textbf{T} (tight loops), and
  \textbf{L} (soft loops). E2B and E4B remain fully clean across all cells. For the
  31B selected variant, the 8192-token / \texttt{rep\_pen=1.15} cell has 0/8 tight
  loops, 4/8 soft loops, and 4/8 natural-EOS endings (vs.\ baseline's 1/8
  natural-EOS at the same cell). For 26B, \texttt{rep\_pen=1.15} increases tight
  loops on this prompt (2/8 to 7/8) but substantially raises natural-EOS
  completions; the two settings optimize different objectives.}
  \label{doom_matrix_fig}
\end{figure}

\FloatBarrier
\subsection{Per-Budget Breakdown and 31B Case Study}

At long budgets, the 31B edit reduces genuine detected loops (excluding the
false-positive \texttt{(none)} bookkeeping described above) from 13/24 to 6/24
(5 tight, 1 soft) at \texttt{rep\_pen=1.0}; counts are essentially identical at
4k and 8k. The decrease in detected loops is offset by increased endless
self-correction, so total doom looping does not fall. Adding
\texttt{rep\_pen=1.15} changes the outcome distribution: tight loops fall to
1/24 at both budgets, and natural-EOS completions rise from 5/24 at
\texttt{rep\_pen=1.0} to 15/24 at 8k. The same penalty on the unedited baseline is
less effective (7/24 detected loops and 12/24 natural-EOS completions at 8k).

For 26B, the edit reduces detected loops from 8--9/24 to 5/24 at
\texttt{rep\_pen=1.0}, of which 3/24 are tight. As with 31B, the reduction in
detected loops is offset by endless self-correction rather than a reduction in
total doom looping. Adding \texttt{rep\_pen=1.15} increases detected loops to
7/24, but raises natural-EOS completions from 3/24 to 15/24 at 4k and from 6/24
to 17/24 at 8k. Thus \texttt{rep\_pen=1.0} minimizes detected loops, whereas
1.15 maximizes natural completion rate.

The \texttt{constellations} prompt illustrates the collateral effect. At 4k tokens
and \texttt{rep\_pen=1.0}, the 31B baseline produces a complete 88-item list in
5/8 seeds. Its reasoning performs an alphabetical-coverage check ending
``\texttt{W:(none), X:(none), Y:(none), Z:(none)}'' before committing. The
selected variant retains this behavior in only 1/8 seeds; on the others, the
closure step disappears and doom looping follows. Adding
\texttt{rep\_pen=1.15} largely closes the gap at 8k, where both variants complete
4/8 seeds. The penalty also changes the baseline's verification behavior, which
is one reason it is recommended for 31B only in conjunction with the outcome
trade-offs reported in the main paper.

\FloatBarrier
\section{Rust Code-Generation Side-Effect Check}
\label{appendix:rust}

The simplest way to reduce the enumeration loops described in main paper Sec.~3 is at sampling time, by raising the repetition penalty (\texttt{rep\_pen}). This is effective on the surface, but repetition penalty is a blunt instrument: it penalizes \emph{any} previously emitted token, including idiomatic Rust macros (\texttt{println!}, \texttt{writeln!}, \texttt{Result}) that a correct long program is expected to reuse. We quantify whether the penalty values that suppress enumeration loops also corrupt unrelated code-generation tasks, and whether this side effect depends on model size. The selected E4B weight edit (\hcode{strip-L18-K3}) operates at \texttt{rep\_pen=1.0} and is therefore expected to bypass this tradeoff; the patched-E4B rows in Table~\ref{tab:rust_sideeffect} are the positive control confirming that it does not introduce a new corruption pathway.

We use a single, deliberately long prompt, \texttt{rust\_clone\_state}: \textit{``Write a complete Rust TUI application that queries Ollama's HTTP API and shows a real-time dashboard with VRAM usage, in-flight inferences, and KV-cache stats. Use crossterm for the TUI. Use \texttt{Arc<Mutex<AppState>>} and clone the state across multiple threads (one polling thread per metric). Include the full \texttt{main.rs} with all imports and the \texttt{AppState} definition.''} The expected output is 3{,}000--4{,}000 tokens of Rust, making the task exercise sustained idiom reuse across a long generation.

For every configuration we run the 30 fixed seeds 0--29 with \texttt{temperature=0.7}, \texttt{top\_p=0.95}, \texttt{enable\_thinking=True}, and \texttt{max\_tokens=4096}. Each completion is classified by two independent signals:
\begin{itemize}
  \item \textbf{Stop reason}: vLLM's \texttt{finish\_reason}. \texttt{stop} means the model emitted the EOS token; \texttt{length} means the model exhausted the 4096-token budget without stopping. Used alone, \texttt{finish\_reason=length} is ambiguous: it may indicate either a coherent answer that exceeded the budget, or a model stuck in non-terminating output.
  \item \textbf{Manual code review}: for every \texttt{length} completion we inspect the tail of the generation. A completion is marked \emph{corrupted} if it contains fabricated macro names (e.g.\ \texttt{writable!}, \texttt{nf!}, \texttt{defer!}, \texttt{stop\_workers!}), fabricated identifiers (\texttt{SysOutMock}, \texttt{MockTerminalInfo}, \texttt{safe\_initialization\_trigger}), incorrectly cased standard items (\texttt{Duration::from\_Millis}), or other clearly non-compiling Rust. Completions whose tails exhibit coherent, idiomatic Rust that simply did not reach EOS within the budget are marked \emph{clean-truncated}.
\end{itemize}
A completion is counted as a \textbf{pass} if \texttt{finish\_reason=stop} \emph{or} the tail review marks it clean-truncated; it is counted as a failure only if it is corrupted.

\paragraph{Results.}
We evaluate the E4B baseline (\texttt{gemma-4-E4B-it}) and the 31B baseline (\texttt{gemma-4-31B-it}) at three repetition penalty values each. Because 31B exhibits no corruption at any penalty value, no patched-31B positive control is warranted; the positive control rows test only the selected E4B weight edit (\hcode{strip-L18-K3}) at the two inference-time values we recommend. All counts are per configuration, $N=30$ (Table~\ref{tab:rust_sideeffect}).

\begin{table}[htbp]
\centering
\small
\renewcommand{\arraystretch}{1.3}
\begin{tabular}{@{}llcccc@{}}
\toprule
\textbf{model} & \textbf{rep\_pen} & \textbf{stop} & \textbf{length} & \textbf{corrupted} & \textbf{pass}\\
\midrule
E4B baseline & 1.00 & 30 & 0  & 0          & \textbf{30 / 30}\\
E4B baseline & 1.15 & 29 & 1  & 0          & \textbf{30 / 30}\\
E4B baseline & 1.30 & 8  & 22 & ${\geq}20$ & \textbf{8 / 30}\\
\midrule
31B baseline & 1.00 & 30 & 0 & 0 & \textbf{30 / 30}\\
31B baseline & 1.15 & 30 & 0 & 0 & \textbf{30 / 30}\\
31B baseline & 1.30 & 30 & 0 & 0 & \textbf{30 / 30}\\
\midrule
E4B \hcode{strip-L18-K3} & 1.00 & 29 & 1 & 0 & \textbf{30 / 30}\\
E4B \hcode{strip-L18-K3} & 1.15 & 30 & 0 & 0 & \textbf{30 / 30}\\
\bottomrule
\end{tabular}
\caption{Rust code-generation outcomes at varying repetition penalties. \emph{Stop} = model emitted EOS naturally; \emph{Length} = model exhausted the 4096-token budget. \emph{Corrupted} counts length-truncated completions confirmed by manual review to contain fabricated macros, fabricated identifiers, or non-compiling syntax (${\geq}20$ at E4B \texttt{rep\_pen=1.30}, with 2 additional borderline cases excluded). \emph{Pass} is the strict count: stop or clean-truncated. The 31B rows show that the corruption pathway is model-specific: at all three penalties 31B produces clean, complete Rust. The bottom two rows confirm that the selected weight edit does not introduce the corruption pathway on E4B.}
\label{tab:rust_sideeffect}
\end{table}

The 31B baseline rows all terminate cleanly at every penalty value (30/30 stop, zero corrupted). However, the picture for the E4B model is more nuanced.
The single \texttt{length} case at E4B \texttt{rep\_pen=1.15} (baseline row 2) and at \texttt{rep\_pen=1.00} (selected variant row 1) are both clean-truncated: the model was writing idiomatic Rust and exhausted the token budget before emitting EOS, thus both count as pass.

At \texttt{rep\_pen=1.30} the picture is qualitatively different for the baseline E4B model. Of the 22 length-truncated completions:
\begin{itemize}
  \item \textbf{1} was caught by the strict periodic-loop heuristic during data collection (seed 15).
  \item A further \textbf{15} contain fabricated macro names or identifiers detectable by a tail regex: \texttt{writable!}, \texttt{nf!}, \texttt{defer!}, \texttt{stop\_workers!}, \texttt{SysOutMock}, \texttt{MockTerminalInfo}, \texttt{safe\_initialization\_trigger}, \texttt{Duration::from\_Millis}, \texttt{application\_printer}, \texttt{crashed!}, \texttt{uxtheme!}.
  \item Manual review of the remaining \textbf{6} yields \textbf{4} additional cases with non-obvious fabrications (invented APIs such as \texttt{Block::fixed\_area}, \texttt{SubArea::with\_vertical\_chunks}, \texttt{simulate\_background\_polling}, \texttt{libc::\_IO\_FILE\_}, \texttt{event::take()}) and \textbf{2} cases that appear clean-truncated in the visible tail.
\end{itemize}
This gives at least 20 corrupted and 2 ambiguous cases. The pass count of 8/30 is the strict count, treating every length-truncated completion as a failure; counting the 2 borderline cases as passes yields 10/30. The qualitative conclusion is the same under either convention.

\paragraph{Sample completions.}
We illustrate the qualitative difference with the final 600 characters of two representative completions on the same \texttt{rust\_clone\_state} prompt.

\medskip
\noindent\textit{Clean completion} (E4B baseline, \texttt{rep\_pen=1.00}, seed 27, \texttt{finish\_reason=stop}, total 2{,}499 tokens):

\begin{quote}\small\ttfamily\raggedright
{\dots}eld, and then releases the lock. It checks the \texttt{is\_running} flag to exit gracefully.\\
4.~\textbf{\texttt{run\_tui}}: The main thread handles the user interface. It runs in a loop, waiting for input (\texttt{crossterm::event::poll}). Crucially, \emph{before} every redraw, it locks the \texttt{Mutex} to get the latest metrics, draws the UI, and then releases the lock.\\
5.~\textbf{Graceful Shutdown}: When the user presses `q', the main thread sets \texttt{state.is\_running = false}. The polling thread detects this flag on its next loop iteration and exits cleanly.
\end{quote}
All macros referenced (\texttt{crossterm::event::poll}, \texttt{Mutex::lock}) are real, the architecture is internally consistent, and the model emitted EOS naturally.

\medskip
\noindent\textit{Corrupted completion} (E4B baseline, \texttt{rep\_pen=1.30}, seed 12, \texttt{finish\_reason=length}, total 4{,}096 tokens):

\begin{quote}\small\ttfamily\raggedright
{\dots}tern crate num\_cpus;\\
external\_crate::protocol::*;\\
\mbox{}\\
enum protocol \{\}\\
trait ProtocolTrait \{\}\\
impl ProtocolTrait for protocol \{\}\\
module tiledb \{ pub trait QueryTermiosMut \{ fn ?(self)\&(mut TcpStream) -> Result<(), AnyHow>(); \}; fn query\_termios(ts: \&mut TcpStream) -> Result<MockTerminalInfo>; \}\\
pub struct MockTerminalInfo
\end{quote}
This is not valid Rust: \texttt{external\_crate} is not a crate keyword, \texttt{module} should be \texttt{mod}, \texttt{fn ?(self)\&(mut TcpStream)} is not valid function syntax, and \texttt{MockTerminalInfo} and \texttt{tiledb} are fabricated identifiers. At this point the model is in a degenerate loop in which each successive attempt to correct the code is itself penalized, causing a drift onto a new fabricated alternative.

\paragraph{Why the corruption appears at rep\_pen=1.30 on E4B but not at 1.15 and not on 31B.}
The standard repetition-penalty processor applies a presence-based transformation
to the current logit of every token that has appeared in the context: positive logits
are divided by the penalty and negative logits are multiplied by it. Repeated
occurrences of the same token do not compound this transformation. A larger penalty
can nevertheless change the ranking between a previously used idiom and unused
alternatives, and can move the idiom outside the \texttt{top\_p=0.95} sampling set.
The E4B outputs are consistent with this pathway: after a common Rust token is
down-ranked, the model samples a fabricated alternative such as \texttt{writable!}.
The absence of the effect in 31B suggests a larger model-specific margin for the
idiomatic continuation, but we did not measure those margins directly. Thus this is
a mechanistic interpretation of the observed corruption, not a causal estimate of
the models' logit margins.

The primary finding of this experiment is that repetition penalty carries a real, model-dependent side effect on code generation: at \texttt{rep\_pen=1.30}, E4B silently corrupts long-form Rust output even while suppressing enumeration loops, whereas 31B, which has a stronger code prior, is unaffected at all three penalty values. The apparent safety of \texttt{rep\_pen=1.15} for E4B should be treated with caution: this result comes from a single prompt, and we did not test settings where idiomatic tokens carry smaller logit margins. A lower corruption threshold may exist for some prompts or models, and we cannot rule it out on the basis of this experiment alone.

The selected weight edit \hcode{strip-L18-K3} sidesteps this observed tradeoff: it achieves loop elimination on E4B at \texttt{rep\_pen=1.0}, and the bottom rows of Table~\ref{tab:rust_sideeffect} show no corruption on this prompt at either \texttt{rep\_pen=1.00} or \texttt{1.15}. Among the values tested here, \texttt{rep\_pen=1.15} produces the better observed trade-off; this single-prompt experiment does not justify a general deployment recommendation.

\FloatBarrier
\section{E2B Intervention}
\label{appendix:e2b_mechanism}

This appendix gives the detailed evidence behind the E2B intervention-layer choice
noted in main paper Sec.~4.

\begin{figure}[htbp]
  \centering
  \includegraphics[width=0.95\textwidth]{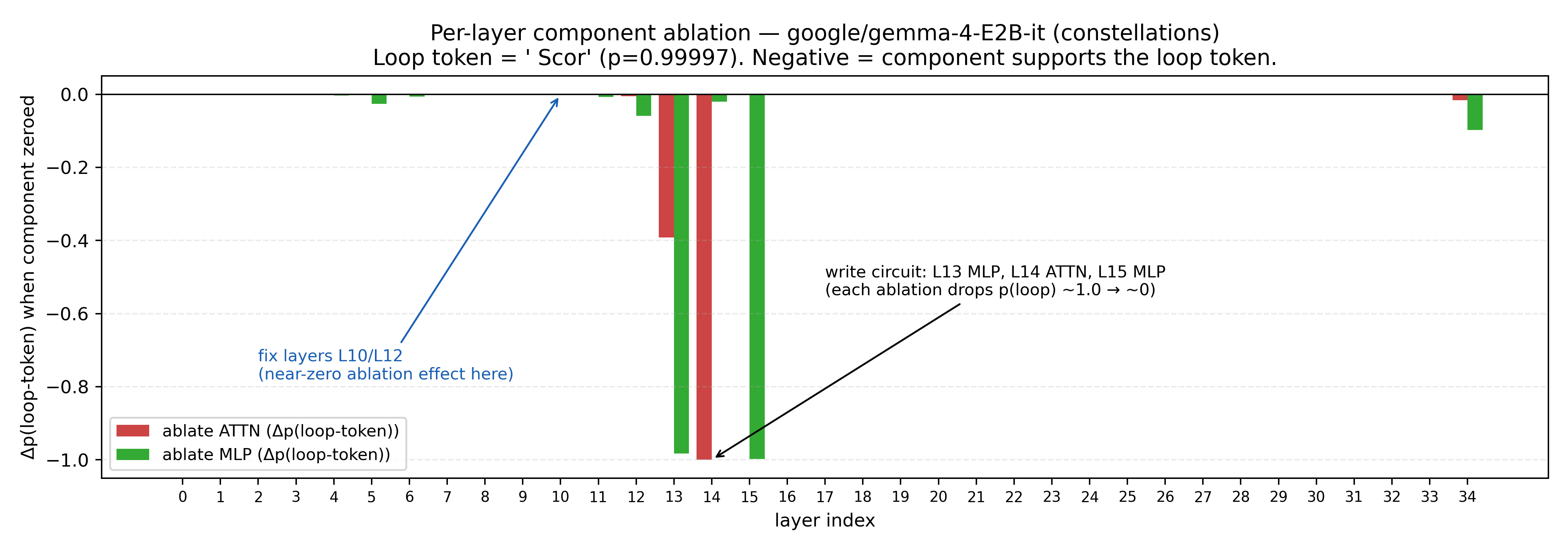}
  \caption{Per-layer component zero-ablation for E2B on the \texttt{constellations}
  loop (loop word `` Scor'', baseline probability $0.99997$). Each bar shows how much
  the probability of the loop word changes when a layer's attention (red) or MLP
  (green) output is zeroed. The word is produced by three consecutive layers
  (L13--L15); the layers we actually edit, L10/L12, show almost no effect here.}
  \label{fig:e2b_ablation}
\end{figure}

On the \texttt{constellations} prompt, the E2B loop repeats one word: at the loop
position the model predicts `` Scor'' (the start of \textit{Scorpius}) with
probability $0.99997$. To find where this word is produced, we zero each layer's
attention or MLP output in turn and measure how much the probability of that word
drops (Figure~\ref{fig:e2b_ablation}). The word is produced by three consecutive
layers: zeroing the L13 MLP drops the probability from $0.99997$ to $0.017$, the L14
attention to almost zero, and the L15 MLP to $0.002$; no other layer matters. For
E4B and 31B, this same test points directly at the layer we edit
(Table~\ref{antiloop_tab}).

E2B is the exception. The neurons we edit are at L10/L12, which come a few layers
before the three layers that produce the word (L13--L15), and there the same
test shows almost no effect. We did not choose L10 from this test; we chose it by
trying edits at every layer from L8 to L20 and measuring the loop rate over full
generations. Stripping pro-loop neurons at one of the later layers (for example L15)
is actually worse than doing nothing: it just makes the model commit to a different
constellation. L10 is the only layer, in our experiments, where editing a few
neurons reliably stops the loop without hurting benchmark scores.

The reason is that a single L10 neuron has only a tiny effect at any one position:
zeroing neuron 3513 at the loop position changes the probability of the loop word by
less than $10^{-4}$, which is why the single-position test above does not flag it.
Its influence is cumulative instead: the neuron's small push toward the loop word,
repeated over the hundreds of tokens the model generates, is what gradually steers
the output into the repeated-constellation answer, and this only becomes visible when
we generate the full response and check whether the loop forms. This is also why the
selected edit \textit{reverses} the neuron's sign rather than simply zeroing it
(main paper Sec.~5.1): zeroing removes the push but lets the model reroute to
the same answer through other neurons, whereas reversing the sign actively steers
away from it. In short, the layers where the loop word \textit{appears}
(L13--L15) are not the layer where it is best \textit{prevented} (L10), similar to the
finding of \citet{hase2023localization} that the place where information is stored in
a model need not be the best place to edit it. We see an analogous gap in 26B, where
the loop signal peaks at different layers than the experts we mask
(Figure~\ref{fig:ablation_26b}).

\begin{table}[htbp]
\centering
\small
\renewcommand{\arraystretch}{1}
\begin{tabularx}{\textwidth}{@{}p{1.2cm}X p{2.2cm}p{2.2cm}p{3.2cm}p{1.8cm}@{}}
\toprule
\textbf{\# neurons} &
\textbf{intervention} &
\textbf{thinking enabled} &
\textbf{thinking disabled} &
\textbf{worst $\Delta$ (benchmark, mode)} &
\textbf{mean $|\Delta|$} \\
\midrule
0 & Baseline & 6/64 & 7/64 & 0.0 pp & 0.00 pp \\
1 & L10:3513 sign inversion ($\alpha=-1$) & 0/64 & 0/64 & $-2.6$ pp (IFEval, disabled) & 0.80 pp \\
\textbf{2} & \textbf{L10:3513 sign inversion ($\alpha=-0.8$) + L12:3838 amplification ($\alpha=3.0$)} & \textbf{0/64} & \textbf{0/64} & \textbf{$-0.74$ pp (IFEval, disabled)} & \textbf{0.56 pp} \\
3 & L10:3513 sign inversion + L11 top-neuron amplification + L12:3838 amplification & 0/64 & 0/64 & $-0.86$ pp (TruthfulQA, enabled) & 0.30 pp \\
4 & Top-3 L10 sign inversion + L12:3838 amplification & 0/64 & 0/64 & $-1.5$ pp (IFEval, disabled) & 0.71 pp \\
6 & Top-3 L10 sign inversion + top-3 L12 amplification & 0/64 & 0/64 & $-3.0$ pp (IFEval, disabled) & 0.87 pp \\
3 & Top-3 L10 stripping & 2/64 & 2/64 & $-2.2$ pp (IFEval, disabled) & 0.76 pp \\
\bottomrule
\end{tabularx}
\caption{E2B interventions grouped by number of modified MLP neurons. Loop counts are reported over the full 8-prompt $\times$ 8-seed $\times$ 2-mode evaluation. Benchmark summaries exclude invalid MMLU results; in the worst $\Delta$ column, ``disabled''/``enabled'' refer to thinking modes.}
\label{tab:e2b_pareto}
\end{table}

\subsection{One-Neuron Held-Out Result}
\label{supp:e2b_one_neuron}

We evaluated the development-grid one-neuron intervention (L10:3513 sign
inversion, $\alpha=-1$) on the same six held-out prompts, unseen seeds 8--15,
and two thinking modes used for the selected variants. Across the 96 paired
generations, baseline$\rightarrow$one-neuron loops are
$6\rightarrow4$: five baseline loops are resolved and three new loops are
introduced (two-sided exact McNemar $p=0.727$). Natural termination changes
from 90/96 to 81/96, and mean unique parsed items from 50.38 to 45.79.
Without thinking, loops change $6/48\rightarrow3/48$ (five resolved, two
introduced; $p=0.453$); with thinking, they change
$0/48\rightarrow1/48$. This result does not provide strong evidence of
held-out generalization for the single-neuron edit, despite its
13/128$\rightarrow$0/128 development-grid result.

\FloatBarrier
\section{Cross-Family and Antidoom Comparisons}
\label{supp:cross_family}

\subsection{Frozen Cross-Family Protocol}

We first ran trigger discovery rather than assuming that Gemma's enumeration
prompts transfer to other families. We screened prompts from Antidoom's released
training mixture~\citep{liquidAI2026Antidoom} by generating outputs from
Qwen3.5-4B~\citep{qwen35modelcard} and
LFM2.5-1.2B-Thinking~\citep{liquidAI2026thinking}, then applying the
repetition detector. This produced 200 detector-positive generations among
1,782 Qwen generations and 27 among 4,992 LFM generations. Prompts were split
before edit selection. Attribution rankings were highly unstable: mean pairwise neuron-rank
Spearman correlation at the selected layer was 0.024 for Qwen and 0.004 for
LFM. As in the Gemma 4 experiments, attribution therefore generates candidates;
full-generation development sweeps select interventions.

The initial 32-prompt development grid selected
\texttt{L12-K30-zero}: it tied \texttt{L12-K100-invert} on loop count and new
loops, then won the frozen tie-break by smaller $K$ and zeroing before
inversion. Refinement used the remaining 108 prompts from the same 140-prompt
development pool and selected \texttt{L12-K100-invert}, reducing detected loops
from 58/108 to 37/108 (30 resolved and 9 introduced). The released package
contains all 35 validation variants and reproduces this selection. After
refinement was complete, we discovered and froze a fresh 400-prompt confirmation set, disjoint
from the development, validation, and earlier 60-prompt held-out splits by
\texttt{split\_key}; none of these prompts was used to select the edit or train
with FTPO. The historical 60-prompt report retains
\texttt{L12-K30-zero}, whereas all confirmation and capability results below
use the refined L12 $K=100$ sign-inversion edit. We evaluated that edit, the
untouched model, and two FTPO checkpoints on identical prompts and decoding.
Table~\ref{tab:qwen_confirmation} reports both our detector and Antidoom's
repeated-character-span detector. As detailed in the blinded audit above, ours
targets short exact token periods and normalized list collapse, while Antidoom
also captures longer repeated character spans.

\begin{table}[htbp]
\centering\small
\begin{tabular}{lrr}
\toprule
\textbf{variant} & \textbf{our detector} & \textbf{Antidoom detector} \\
\midrule
Baseline & 74 & 181 \\
Sparse L12 $K=100$ inversion & 42 (56/24) & 104 (111/34) \\
FTPO, early stopped & 0 (74/0) & 0 (181/0) \\
FTPO, full epoch & 23 (68/17) & 26 (165/10) \\
\bottomrule
\end{tabular}
\caption{Qwen3.5 frozen confirmation results ($n=400$). Parentheses give
resolved/introduced positives relative to baseline.}
\label{tab:qwen_confirmation}
\end{table}

For the sparse edit, the exact paired $p$-values are 0.00045 under our detector
and $9.9\times10^{-11}$ under the Antidoom detector. The corresponding
early-stopped FTPO values are $1.1\times10^{-22}$ and
$6.5\times10^{-55}$. These tests establish changes under each operational
detector; without human labels, they do not establish that one detector is more
valid. On the 400 baseline outputs, 71 are positive under both detectors, 110
under Antidoom only, 3 under ours only, and 216 under neither.
Cases detected under Antidoom only are generally long-range terminal repetition (median period
322 characters), whereas our detector additionally targets normalized
list-collapse repetition.

\paragraph{Antidoom-detector termination audit.}
Detector-negative cases do not necessarily mean that a generation completed.
The Antidoom detector tells us whether repetition remains; vLLM's finish reason
tells us whether the response ended naturally or reached the token limit. The
untouched Qwen3.5-4B baseline has 181/400 detected loops and 95/400 natural
stops. All loop counts in this paragraph use the Antidoom detector. The sparse edit
removes detected repetition from 111 baseline loops: 57 then stop naturally,
while 54 still reach the token limit. Early-stopped FTPO removes detected
repetition from all 181 baseline loops, but only 68 stop naturally and 113 still
reach the limit. Full-epoch FTPO resolves 165 baseline loops; 48 stop and 117
reach the limit. Thus a detector-negative output is not necessarily a completed
answer: it may be productive but truncated, or it may continue non-periodic
self-correction.

\subsection{Capability and Training Comparison}

We evaluate Qwen on the eight benchmarks listed in
Table~\ref{tab:qwen_cap}, with thinking enabled. We exclude MMLU because its
short response cap truncates prose-first responses. The table
shows that the early-stopped FTPO checkpoint remains close to baseline across
these eight tasks, while the full-epoch checkpoint is substantially
degraded on six. Sparse editing is also close to or above baseline on this
set of benchmarks, but these finite tests cannot establish preservation on unmeasured
behavior.

\begin{table*}[htbp]
\centering\small
\begin{tabular}{lrrrrrrr}
\toprule
\textbf{variant} & \textbf{IFEval} & \textbf{ARC-E} & \textbf{ARC-C} &
\textbf{GSM8K} & \textbf{BoolQ} & \textbf{HellaSwag} & \textbf{WinoGrande} \\
\midrule
Baseline & .277 & .996 & .956 & .882 & .778 & .728 & .804 \\
Sparse edit & .281 & .992 & .962 & .952 & .842 & .770 & .866 \\
FTPO-early & .281 & .994 & .956 & .898 & .870 & .736 & .838 \\
FTPO-full & .257 & .644 & .603 & .154 & .554 & .434 & .310 \\
\bottomrule
\end{tabular}
\caption{Qwen3.5 capability scores. FTPO-early is the checkpoint selected by
the pre-specified chosen-win stopping rule; FTPO-full disables that rule.}
\label{tab:qwen_cap}
\end{table*}

We generated 20,000 unique Qwen FTPO pairs; 10,777
examples remained after trainer-side filtering. Both Qwen runs use the same
LoRA configuration, effective batch size 16, and at most one epoch; only the
chosen-win early-stopping threshold differs. Thus the full-epoch result is an
overtraining control rather than a separately tuned method.

The same pipeline is less effective on E4B. Generation yielded 2,080 pairs
(2,076 after filtering), and the matched held-out primary endpoint remains
$18/96\rightarrow18/96$: four baseline loops resolve and four new loops appear
(exact McNemar $p=1$). This contrast shows that FTPO's benefit is
model/data-yield dependent rather than automatic.

\subsection{LFM Boundary Result}

For LFM2.5-1.2B-Thinking, L14 $K=100$ zeroing changes development loops from
4/18 to 1/18 and held-out loops from 4/9 to 1/9 (three resolved, none
introduced; exact McNemar $p=0.25$). Five matched random masks produce one to
four loops, including one tie with the selected edit. Capability deltas range
from $-6.4$ pp on WinoGrande to $+1.0$ pp on ARC-Challenge. We therefore treat
LFM as sparse-edit feasibility only, not evidence of localization specificity
or capability preservation.

\FloatBarrier
\section{Implementation Identifiers and Model Revisions}
\label{appendix:implementation}

Table~\ref{variant_names_tab} maps the readable names used in the paper to exact
experiment identifiers. Table~\ref{tab:hf_revisions} records the pinned base-model
revisions.

\begin{table}[htbp]
\centering\small
\begin{tabularx}{\textwidth}{@{}l l X X@{}}
\toprule
\textbf{model} & \textbf{paper ID} & \textbf{internal ID} & \textbf{description} \\
\midrule
E2B & \texttt{E2B-2N} & \texttt{flipL10K1-a-0.8 + ampL12K1-a3.0} &
Two-neuron sign inversion and amplification \\
E4B & \texttt{E4B-L18-3N} & \texttt{strip-L18-K3} &
Three-neuron L18 strip \\
31B & \texttt{31B-L36-1100N} & \texttt{strip-ff-K1000-wonly-K100} &
Two-pass L36 strip (1,100 selections; 1,062 unique neurons) \\
26B & \texttt{26B-3E} & \texttt{bake-mask-v2\_top3} &
Three-expert routing mask \\
Qwen & \texttt{Qwen-L12-100N} & \texttt{L12-K100-invert} &
L12 sign inversion of 100 selected neurons \\
LFM & \texttt{LFM-L14-100N} & \texttt{L14-K100-zero} &
L14 zeroing of 100 selected neurons \\
Qwen & \texttt{FTPO-early} & \texttt{primary-earlystop-r3} &
Antidoom FTPO checkpoint selected by early stopping \\
Qwen & \texttt{FTPO-full} & \texttt{full-epoch-r1} &
Antidoom FTPO one-epoch control checkpoint \\
\bottomrule
\end{tabularx}
\caption{Selected intervention identifiers.}
\label{variant_names_tab}
\end{table}

\subsection{Hugging Face Model Revisions}
\label{appendix:hf_revisions}

For reproducibility, Table~\ref{tab:hf_revisions} lists the exact Hugging Face
commits of the six base models. All are loaded as
\texttt{AutoModelForCausalLM.from\_pretrained(model\_id, revision=<sha>)}.

\begin{table}[htbp]
\centering
\small
\begin{tabular}{lll}
\toprule
\textbf{Short name} & \textbf{Hugging Face model ID} & \textbf{Revision (commit SHA)} \\
\midrule
E2B     & \texttt{google/gemma-4-E2B-it}     & \texttt{905e84b50c4d2a365ebde34e685027578e6728db} \\
E4B     & \texttt{google/gemma-4-E4B-it}     & \texttt{d6436b3d62967e1af08bbb046c6300b2a9ae8e85} \\
26B & \texttt{google/gemma-4-26B-A4B-it} & \texttt{6e6f6edea8c52db2094dca3086e4b963a0034dfc} \\
31B     & \texttt{google/gemma-4-31B-it}     & \texttt{fb9ae262347c3945692f09a612f8bb189def854f} \\
Qwen & \texttt{Qwen/Qwen3.5-4B} & \texttt{851bf6e806efd8d0a36b00ddf55e13ccb7b8cd0a} \\
LFM & \texttt{LiquidAI/LFM2.5-1.2B-Thinking} & \texttt{95053d21d8e0b7ca99421a2127ae39c64f685ff3} \\
\bottomrule
\end{tabular}
\caption{Pinned Hugging Face revisions of the six base models. The four Gemma
commits are the revisions used for the long-budget re-sweeps reported in the
main paper. Earlier Gemma attribution and intervention sweeps used revisions
with byte-identical safetensors; only \texttt{chat\_template.jinja} differed:
\url{3555bddc93a623db8887dd2e52123facc45ade77} (E4B),
\url{462a98a12e28e2cbcfccaf78fe41e3e50235e6ae} (26B-A4B), and
\url{ba74f5b6c647c0911554e50278d6f6f4477f9010} (31B). E2B received no such
update. The changed template branch handles multimodal placeholders inside
tool-response blocks and is never entered by our text-only probes. All sweeps
load model and tokenizer files from the same local Hugging Face cache rather
than the live Hub.}
\label{tab:hf_revisions}
\end{table}

\section{Reproducibility Protocol}
\label{appendix:reproducibility}

\subsection{End-to-End Procedure}
The following procedure summarizes the experimental method and separates
development from frozen evaluation.
\begin{enumerate}
\item Generate complete responses to the development probes and classify their
tails with the fixed tight-loop and normalized-list-collapse detector.
\item For detector-positive responses, zero one layer or component at a time,
then rank neurons (or routed experts) using the attribution score defined in
Sec.~4 of the main paper.
\item Construct candidate edits by stripping, sign-inverting, or amplifying the
ranked units; for the routed model, mask candidate experts.
\item Sweep only the declared development layers, edit sizes, and operations.
Select the smallest candidate on the loop-rate/capability Pareto frontier; no
held-out or confirmation response is used for selection.
\item Freeze the edit specification, prompts, decoding settings, model
revision, and seeds. Evaluate the untouched and edited models once on every
prompt--seed--thinking-mode cell, then compute paired loop, termination,
coverage, benchmark, and random-control statistics.
\end{enumerate}

\subsection{Randomness, Runs, and Development Ranges}
Unless explicitly stated otherwise, every reported cell contains one sampled
generation for each prompt--seed--thinking-mode combination; we do not average
repeated runs of the same cell. Gemma development uses seeds 0--7 and held-out
evaluation uses unseen seeds 8--15. The Rust side-effect study uses seeds
0--29. Cross-family discovery and FTPO use seed 3407; sparse-edit development
uses seeds 3407--3410 and held-out evaluation uses 3407--3414. The five matched
random-control masks are drawn once with seed 27012027.

Development searches use E2B layers 8--20; E4B edit sizes including
$K\in\{1,3,10\}$ at the selected layer; 31B prompt-specific and combined
candidate sizes reported in Table~\ref{31b_comb_tab}; and 26B expert masks of
sizes 3, 5, and 10. Cross-family neuron sweeps use the two best development
layers, operations \{zero, sign inversion\}, and
$K\in\{1,3,10,30,100\}$. Candidates are selected first by development loop
count and then by the smallest measured worst-case benchmark regression.

\subsection{Final Generation and Training Settings}
Gemma initial and held-out evaluations use temperature 0.7,
\texttt{top\_p=0.95}, \texttt{top\_k=64}, no repetition penalty, and 1,536
new tokens; long-budget runs use 8,192 new tokens. Cross-family discovery,
sparse-edit evaluation, and FTPO pair generation use temperature 0.01,
\texttt{top\_p=1.0}, \texttt{top\_k=50}, \texttt{min\_p=0.01}, and up to
4,000 new tokens. Capability evaluations use deterministic decoding.

Qwen FTPO trains at most 12,000 filtered examples from the 20,000 generated
pairs for one epoch, with sequence length 6,000, learning rate
$1.5\times10^{-5}$, warmup ratio 0.1, weight decay 0.01, effective batch size
16, and bfloat16. LoRA uses rank 256, scale 128, zero dropout, and the final
five transformer layers, targeting attention projections, MLP projections,
and the language-model head. The selected checkpoint uses a chosen-win
early-stopping threshold of 0.4; the full-epoch control disables practical
early stopping with threshold 1.1.

\subsection{Compute and Software}
Experiments ran on Linux Slurm workers with one NVIDIA H100 80\,GB GPU per
job. The Hugging Face reference environment used Python 3.10.12,
PyTorch 2.11.0 with CUDA 12.8, and Transformers 5.8.0. The Antidoom/FTPO
environment used Python 3.12.13, PyTorch 2.11.0 with CUDA 13.0,
Transformers 5.6.2, PEFT 0.19.1, TRL 1.2.0, and vLLM 0.20.1. Inference and
training used bfloat16. The retained capability-evaluation environment uses
Inspect-AI 0.3.220 and Inspect-Evals 0.5.1; the submitted requirements files
record these software versions.

\FloatBarrier

\end{document}